\documentclass{article}
\usepackage[final]{corl_2019} % Uncomment for the camera-ready ``final'' version
\usepackage{graphicx} % for pdf, bitmapped graphics files
\usepackage{wrapfig}
\usepackage{amsmath}
\usepackage{amsfonts}
\usepackage{amssymb}
\usepackage{bbm}
\usepackage{xcolor}
\usepackage[linesnumbered,ruled,vlined]{algorithm2e}
\SetKwInput{KwInput}{Input}                % Set the Input
\SetKwInput{KwOutput}{Output}              % set the Output

\usepackage{url}

\title{Variational Inference MPC for \\ Bayesian Model-based Reinforcement Learning}

\author{
  Masashi Okada \\
  Panasonic Corp., Japan \\
  \texttt{okada.masashi001@jp.panasonic.com} \\
  \And
  Tadahiro Taniguchi\\
  Ritsumeikan Univ. \& Panasonic Corp., Japan \\
  \texttt{taniguchi@em.ci.ritsumei.ac.jp}  \\
}

\newcommand{\optimal}{\mathcal{O}}

\newcommand{\traj}{\tau}
\newcommand{\state}{\boldsymbol{s}}
\newcommand{\action}{\boldsymbol{a}}
\newcommand{\actionm}{\boldsymbol\mu}
\newcommand{\actions}{\boldsymbol\Sigma}
\newcommand{\mixcoef}{\pi}
\newcommand{\resp}{\eta}
\newcommand{\modelpost}{p_{\mathcal{D}}(\theta)}
\newcommand{\qa}{q_{\action}}
\newcommand{\qt}{q_{\traj}}
\newcommand{\proposedmethod}{PaETS}{}
\newcommand{\totalreward}{r(\traj)}
\newcommand{\figref}[1]{Fig.~\ref{#1}}
\newcommand{\tabref}[1]{Table~\ref{#1}}
\newcommand{\secref}[1]{Sec.~\ref{#1}}

\begin{document}
\maketitle

%===============================================================================

\begin{abstract}
% The purpose of this document is to provide both the basic paper template and submission guidelines.
% Abstracts should be a single paragraph, between 4--6 sentences long, ideally. Gross violations will trigger corrections at the camera-ready phase.
In recent studies on model-based reinforcement learning (MBRL), incorporating uncertainty in forward dynamics is a state-of-the-art strategy to enhance learning performance, making MBRLs competitive to cutting-edge model-free methods, especially in simulated robotics tasks. %, like Soft Actor-Critic.
Probabilistic ensembles with trajectory sampling (PETS) is a leading type of MBRL, which employs Bayesian inference to dynamics modeling and model predictive control (MPC) with stochastic optimization via the cross entropy method (CEM).
In this paper, we propose a novel extension to the uncertainty-aware MBRL.
Our main contributions are twofold:
Firstly, we introduce a variational inference MPC (VI-MPC), which reformulates various stochastic methods, including CEM, in a Bayesian fashion.
Secondly, we propose a novel instance of the framework, called probabilistic action ensembles with trajectory sampling (\proposedmethod).
As a result, our Bayesian MBRL can involve multimodal uncertainties both in dynamics and optimal trajectories.
In comparison to PETS, our method consistently improves asymptotic performance on several challenging locomotion tasks.
% However, CEM and other similar stochastic methods are not uncertainty-aware and tend to underestimate the multimodal uncertainty in optimal trajectories, ignoring precious information in it.
% In order to develop uncertainty-aware optimization, this paper first introduce a variational inference MPC, which reformulates the stochastic methods as a Bayesian fashion.
% In addition,
\end{abstract}

\keywords{model predictive control, variational inference, model-based reinforcement learning}

%===============================================================================

\section{Introduction}
Model predictive control (MPC) is a powerful and accepted technology for advanced control systems such as %
manufacturing processes~\citep{vargas2000multilayer}, %qin2003survey,
HVAC systems~\citep{afram2014theory},
power electronics~\citep{vazquez2014model},
autonomous vehicles~\citep{paden2016survey}, %,kamel2017model
and humanoids~\citep{kuindersma2016optimization}. %erez2013integrated
MPC utilizes the specified models of system dynamics to predict future states and rewards (or costs)
to plan future actions that maximize the total reward over the predicted trajectories.
Especially for industrial applications, the clear explainability of such a decision-making process is advantageous.
Furthermore, in some tasks (e.g.,~games)~\citep{silver2016mastering}, planning-based policies of this nature could outperform \textit{reactive}-policies (e.g.,~full neural network policies).

Model-based reinforcement learning (MBRL) methods that employ expressive function approximators (e.g.,~deep neural networks: DNNs)~\citep{deisenroth2011pilco,williams2017information,nagabandi2018neural} present appealing approaches for MPC.
The main difficulty in introducing MPC to practical systems is specifying the forward dynamics models of target systems.
However, accurate system identification is challenging in many advanced applications.
Take robotics for example, where robots encounter floors and walls, and must be able to manipulate some objects, making the dynamics highly non-linear.
% An appealing approach to this is to employ expressive function approximators (e.g.,~deep neural networks: DNNs) and
% model-based reinforcement learning (MBRL) methods~\citep{deisenroth2011pilco,williams2017information,nagabandi2018neural}.
The main objective of MBRL is to train approximators of complex dynamics through experiences in real systems.
The general procedure of MBRL is summarized as;
(1.~\textit{training-step}) train the approximate model with a given training dataset, then
(2.~\textit{test-step}) execute the actions (or policies) optimized with the dynamics model in a real environment and augment the dataset with the observed results.
The above training and test steps are iteratively conducted to collect sufficient and diverse data so as to achieve the desired performance.

One feature of MBRL is its considerable sample efficiency compared to model-free reinforcement learning (MFRL), which directly trains policies through experiences.
In other words, MBRL requires much less test time in real environments.
In addition, MBRL benefits from the generalizability of the trained model, which can be easily applied to new tasks in the same system.
However, the asymptotic performance of MBRL is generally inferior to that of model-free methods.
This discrepancy is primarily due to the overfitting of dynamics models to the few data available during initial MBRL steps,
which is called the \textit{model-bias problem}~\citep{deisenroth2011pilco}.
Several studies have demonstrated that incorporating uncertainty in dynamics models can alleviate this issue.
The uncertainty-aware modeling is realized by Bayesian inference employing a Gaussian Process~\citep{deisenroth2011pilco}, \textit{dropout as variational inference} \citep{gal2016dropout,gal2017concrete,kahn2017uncertainty}, or neural network ensembles~\citep{chua2018deep,kurutach2018model,clavera2018model}.

% Especially for initial steps of MBRL, where only a few data is available,
% the trained dynamics model is unreliable and optimized actions would be sub-optimal.
% Executing such the sub-optimal actions may result in undesired behaviors
% so we cannot expect that these newly collected trajectory data will not contribute to enhance the dynamics model.
% By iterating the above steps, dynamics and optimized actions prone to be \textit{biased} (or \textit{overfitted}),
% which is called \textit{model-bias}~\citep{deisenroth2011pilco}.

Probabilistic ensembles with trajectory sampling (PETS)~\citep{chua2018deep} is one type of uncertainty-aware MBRL.
% For instance, an MPC-oriented MBRL method called PETS (probabilistic ensembles with trajectory sampling) has been proposed.
As an MPC-oriented MBRL method, PETS conducts trajectory optimization via the cross entropy method (CEM)~\citep{botev2013cross}
by using trajectories probabilistically sampled from the ensemble networks.
Experiments have demonstrated that PETS can achieve competitive performance over state-of-the-art MFRL methods like Soft Actor Critic (SAC)~\citep{haarnoja2018soft}, while yielding much higher sample efficiency.
Since our primary interest is MPC and its application to practical systems, this paper mainly focuses on PETS and treats this method as a strong baseline.

Considering the success of probabilistic dynamics modeling,
incorporating uncertainty in optimal trajectories appears very promising for MBRL.
However, an optimization scheme that can utilize uncertainty has not yet been discussed.
Although several stochastic approaches,
including CEM,
model predictive path integral (MPPI)~\citep{williams2016aggressive,williams2017information},
covariance matrix adaptation evolution strategy (CMA-ES)~\citep{hansen2003reducing},
and proportional CEM (Prop-CEM)~\citep{goschin2013cross},
have been proposed, they are not uncertainty-aware and tend to underestimate uncertainty.
% Considering the success of uncertainty-aware dynamics modeling, incorporating the uncertainties in optimalities would be very promising.
In addition, although their optimization procedures are very similar, they have been independently derived.
Consequently, theoretical relations among these methods are unclear, preventing us from systematically understanding and reformulating them to be uncertainty-aware in a Bayesian fashion.

\begin{figure}[t]
    \centering
    \begin{minipage}{0.45\textwidth}
        \centering
        \includegraphics[width=\textwidth]{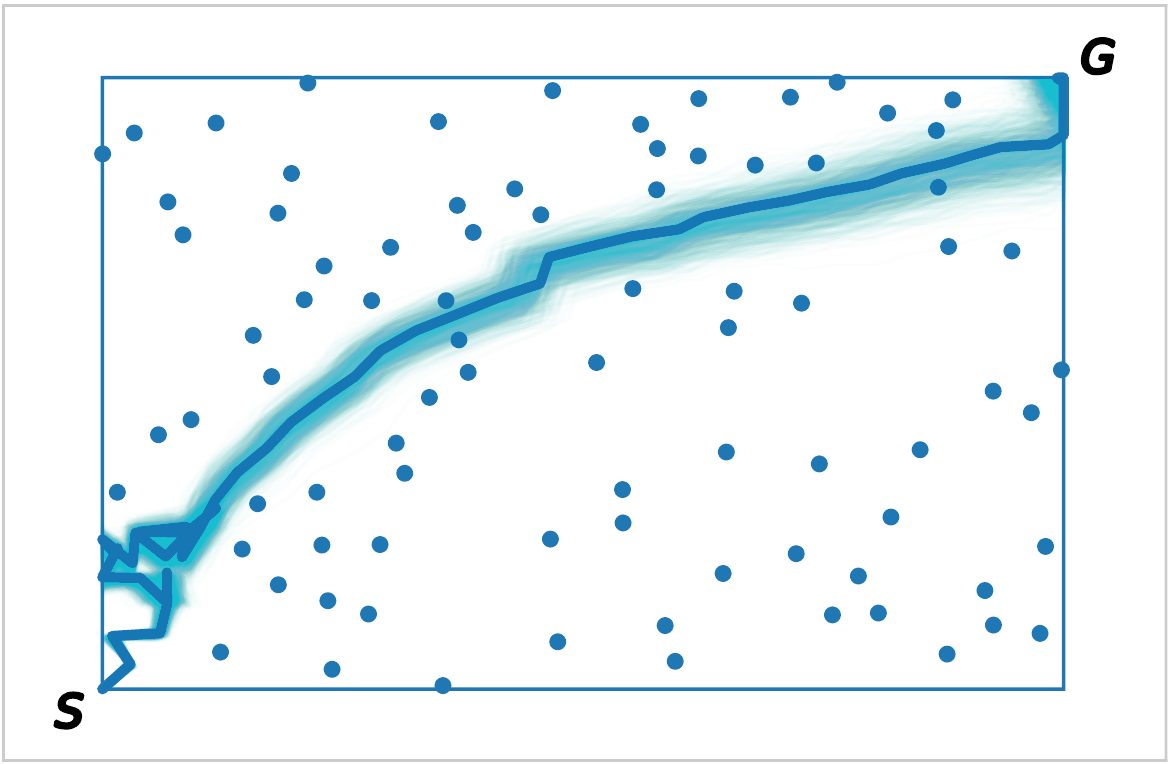}
        % \begin{flushleft}
          {\small (a) Vanilla CEM used in PETS~\citep{chua2018deep}:} \\ {\footnotesize \texttt{VIMPC(`CEM', `Gaussian', False)}}
        % \end{flushleft}
    \end{minipage}
    \hspace{0.05\textwidth}
    \begin{minipage}{0.45\textwidth}
        \centering
        \includegraphics[width=\textwidth]{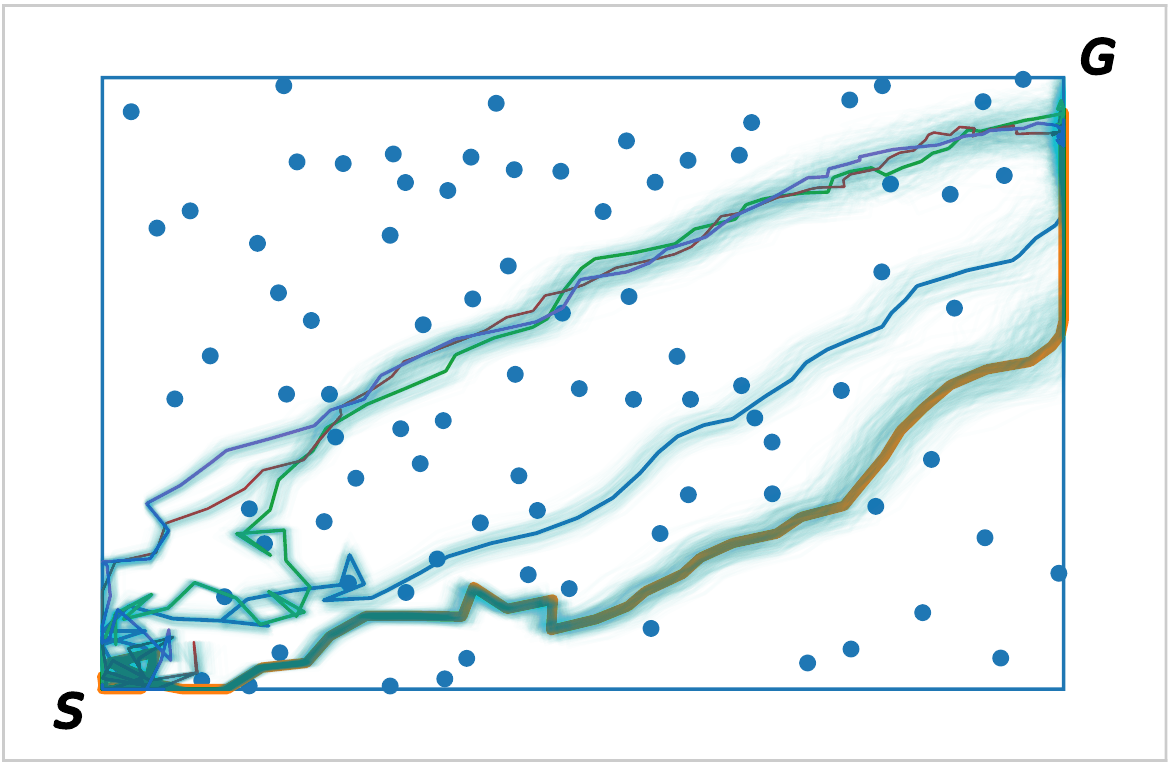}
        % \begin{flushleft}
          {\small (b) \proposedmethod{} (Ours): \\ {\footnotesize \texttt{VIMPC(`CEM', `GMM(M=5)', True)}}}
        % \end{flushleft}
    \end{minipage}
    \caption{
    Toy task examples that illustrate the concept of our method.
    The objective of this task is to navigate a point mass on the $x$-$ y$ plane
    % from $S$ to $G$
    by actuating $\action_{t} = (\Delta x, \Delta y)$ with maximum magnitude $||\action_{t}|| = 0.05$, while avoiding obstacles $\bullet$. %denoted with
    This task is designed to have multiple (sub-)optimal trajectories.
    (a) A trajectory found by vanilla CEM.
    % Since CEM assumes single Gaussian distribution to approximate action posterior, only single trajectory was found.
    (b) Multiple trajectories found by \proposedmethod{} that approximates the trajectory posterior via variational inference with a Gaussian mixture model.
    The line-width indicates the magnitude of mixture-coefficients.
    Exploiting diverse plans encourages active exploration in state-action spaces, improving the optimization performance and training dataset diversity.
    The notation of \texttt{VIMPC()} is introduced in \secref{sec:from_mm_to_vi}.
    }
    \label{fig:toytask}
\end{figure}

Motivated by these, in this paper, we propose a novel MPC concept for Bayesian MBRL.
The organization and contributions of this paper are summarized as follows.
(1)~In \secref{sec:from_mm_to_vi}, we introduce a novel MPC framework, variational inference MPC (VI-MPC), which generalizes and reformulates various stochastic MPC methods in a Bayesian fashion.
The key observations for deriving this framework are organized in \secref{sec:mbrl_review},
where we point out that general stochastic optimization methods can be regarded as the moment matching of the optimal trajectory posterior, which appear in a Bayesian MBRL formulation.
% firstly reformulate MBRL problem as \textit{fully Bayesian} by introducing \textit{control as inference} framework~\citep{levine2018reinforcement}.
% Then, we point out that various stochastic MPC methods can be regarded as the moment matching of the optimal action posterior.
% Based on this observation, we introduce variational inference to approximate the posteior, deriving a novel MPC framework, variational inference MPC, which generalizes and extends the preivous methods to be Bayesian.
(2)~In \secref{sec:vimpc}, we propose a novel instance of the framework, called probabilistic action ensembles with trajectory sampling (\proposedmethod{}).
Toy task examples and the concept of our method are exhibited in \figref{fig:toytask}.
(3)~In \secref{sec:experiment}, we demonstrate that our method consistently outperforms PETS via experiments with challenging locomotion tasks in the MuJoCo physics simulator~\citep{todorov2012mujoco}.

\section{Model-based Reinforcement Learning as Bayesian Inference} \label{sec:mbrl_review}
\begin{wrapfigure}{r}{0.35\textwidth}
    \centering
    \includegraphics[width=0.3\textwidth]{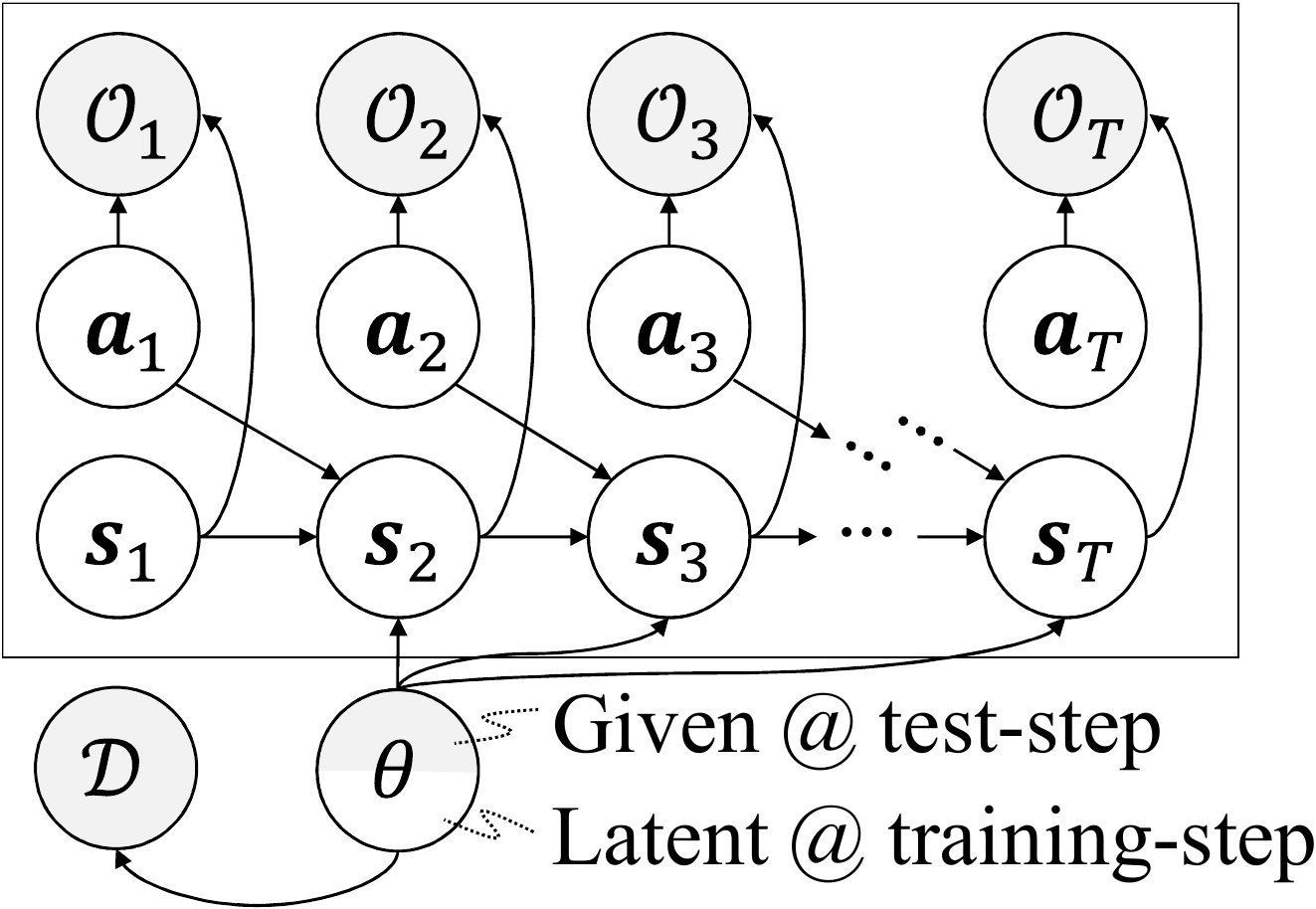}
    \caption{Graphical model for Bayesian MBRL.}
    \label{fig:graphical_model}
\end{wrapfigure}

In this section, we describe MBRL as a Bayesian inference problem using \textit{control as inference} framework~\citep{levine2018reinforcement}.
\figref{fig:graphical_model} displays the graphical model for the formulation,
with which an MBRL procedure can be re-written in a Bayesian fashion:
% therefore we need solve this problem with two steps:
(1.~\textit{training-step}) do inference of $p(\theta|\mathcal{D})$. (2.~\textit{test-step}) do inference of $p(\traj |\optimal_{1:T}=\mathbf{1})$, then, sample actions from the posterior and execute the actions in a real environment.
We denote a trajectory as $\traj := \{(\state_{t}, \action_{t})\}_{t=1}^{T}$, where
$\state_{t}$ and $\action_{t}$ respectively represent state and action.
Given a state-action pair at time $t$, the next state can be predicted by a forward-dynamics model $\state_{t+1} \sim p(\state_{t+1}|\state_{t}, \action_{t}, \theta)$ parameterized with $\theta$.
The posterior of $\theta$ is inferred from training dataset $\mathcal{D}$,
where $\mathcal{D} = \{(\state_{t}, \action_{t}, \state_{t+1})\}$ consists of states and actions \textit{observed} during the test step.
To formulate optimal control as inference, we auxiliarly introduce a binary random variable $\mathcal{O}_{t} \in \{0, 1\}$
to represent the \textit{optimality} of ($\state_{t}$, $\action_{t}$). %\footnote{
% In general, the probabilty of $\optimality$ is defined as $p(\optimality|\state_{t}, \action_{t}) = \exp(r(\state_{t}, \action_{t}))$ \cite{chua2018deep,piche2018probabilistic}.
% }
%
Given $p(\theta|\mathcal{D})$, trajectory optimization can be expressed as an inference problem:
\begin{equation}
    p(\traj |\optimal) \propto
    \int
    \underbrace{
      \left\{
      \prod^{T}_{t=1} p(\optimal_{t}=1| \state_{t}, \action_{t})\right\}
    }_{:= p(\optimal| \traj)}
    \cdot
    \underbrace{
      p\left(\state_{1}\right) \left\{\prod_{t=1}^{T} p\left(\state_{t+1} | \state_{t}, \action_{t}, \theta \right)\right\}
    }_{:= p(\state|\action, \theta)}
    \cdot
    \underbrace{
      p(\theta|\mathcal{D})
    }_{:= \modelpost}
    %
    %  p(\mathcal{D})
        %
    % \cdot
    %
    d\theta \label{eqn:traj_posterior},
\end{equation}
where uninformative action prior (i.e.,~$p(\action_{t}) =\mathcal{U}$: uniform distribution) is supposed. For readability, $\mathcal{O}_{1:T} = \mathbf{1}$ is simply denoted as $\mathcal{O}$.
For the same reason, we omit the subscripts of sequences $\action_{1:T}$, $\state_{1:T}$.
In the remainder of the paper, this simplified notation is employed.
% Eq.~\eqref{eqn:traj_posterior} is written in hierarchical form.
In \secref{sec:dynamics_inference}--\ref{sec:control_inference}, we review how these inference problems have been approximately handled in previous works.
%
% Model-based approaches tend to rely on accurate (learned) dynamics models to solve a task. If
% the dynamics model is not sufficiently precise, the policy optimization is prone to overfit on the
% deficiencies of the model, leading to sub-optimal behavior or even to catastrophic failures. This
% problem is known in the literature as model-bias~\citep{deisenroth2011pilco}.

\subsection{Inference of Forward-dynamics Posterior $\modelpost$} \label{sec:dynamics_inference}
Given a sufficiently parameterized expressive model, i.e., DNNs,
one of the most practical and promising schemes for approximating the posterior $\modelpost$ is to utilize neural network ensembles~\citep{chua2018deep,kurutach2018model,clavera2018model}.
This process approximates the posterior as a set of \textit{particles} $\modelpost \simeq \frac{1}{E}\sum^{E}_{i}\delta(\theta - \theta_{i})$,
where $\delta$ is Dirac delta function and $E$ is the number of networks.
Each particle $\theta_{i}$ is independently trained by stochastic gradient descent so as to (sub-)optimize
$\log \modelpost \propto \log p(\mathcal{D}|\theta)p(\theta)$.
Although this approximation is incompletely Bayesian, this scheme has several useful features.
First, we can simply implement this process in standard deep learning frameworks.
Furthermore, the ensemble model successfully involves multimodal uncertainty in the exact posterior.
% Stein Variational Gradient Descent (SVGD)~\citep{liu2016stein} is an extention of the ensemble approach.
% By introducing kernel method, a particle update law can be theoretically derived,
% in which particles are updated with standard SGD but the particles influences each other by mutual repulsive forces.
% However, SVGD experiments and applications are limited to shallow neural networks so far.

Another possible way to infer $\modelpost$ is \textit{dropout as variational inference}~\citep{gal2016dropout,gal2017concrete,kahn2017uncertainty},
which approximates $\modelpost$ as a Gaussian distribution $q(\theta)$.
It is proofed that the variational inference problem: $\operatorname{argmin}_{q}\operatorname{KL}(q(\theta)||\modelpost)$ approximately equivalent to training networks with dropout, where $\operatorname{KL}(\cdot||\cdot)$ denote Kullback-Leibler (KL) divergence.
Although this scheme is also simple and theoretically supported,
approximation by a single Gaussian distribution tends to underestimate uncertainty (or multimodality) in the posterior.
To remedy this problem, $\alpha$-divergence dropout has been proposed~\citep{li2017dropout}, which replaces KL-divergence to $\alpha$-divergence so as to prevent $q(\theta)$ from overfitting a single mode.
However, as long as $q(\theta)$ is Gaussian, the multimodality cannot be managed well.
% compared to the ensemble scheme.

In our preliminary experiments of MBRL, we have tested the above two schemes and observed that the ensemble performs much better than ($\alpha$-)dropout (this result is summarized in \secref{sec:preliminary}).
This result provides us with the insight that capturing multimodality in the posterior has crucial effects in MBRL literature.
Therefore, in this paper, we also employ this ensemble scheme to approximate $\modelpost$ in the same way as our baseline: PETS~\citep{chua2018deep}.
In \secref{sec:vimpc}, we also attempt to incorporate multimodality in the posterior $p(\traj|\mathcal{O})$.

\subsection{Moment Matching of Trajectory Posterior $p(\traj|\mathcal{O})$} \label{sec:control_inference}
This section clarifies the connection between trajectory optimization and the posterior approximation problem.
The key observation delineated here is that several MPC methods, including CEM used in PETS and MPPI, can be regarded as the moment matching of the posterior.
% Ref~\citep{chua2018deep} introduce cross entropy method (CEM)~\citep{botev2013cross} to solve optimal control problem utilizing the inferred prediction model posterior $\modelpost$.
% In this section, we point out that this optimization can be considered

Given an inferred model posterior $\modelpost$, we can sample trajectories from \eqref{eqn:traj_posterior}.%
\footnote{
Trajectory sampling methods with $\modelpost$ have been discussed and experimented in \citep{chua2018deep}.
In this paper, we employ the TS1 method suggested in the reference (see $\ell$3--6 in Alg.~\ref{alg:vi_traj_opt}).
% , where next state $\state^{(i)}_{t+1}$ in a certain trajectory $i$ is sampled with a time-varying model parameter:
% $\state^{(i)}_{t+1} \sim p(\state_{t+1} | \state^{(i)}_{t}, \action_{t}, \theta^{(i)}_{t}), \theta^{(i)}_{t} \sim \modelpost$
}
Let us approximate the action posterior with a Gaussian distribution $q(\action; \actionm, \actions)$.
The mean of posterior action sequence $\actionm$ can be estimated by moment matching:
\begin{equation}
\actionm =
\mathbb{E}\left[\action \cdot p(\traj|\mathcal{O})\right]
=
 \frac
 {\mathbb{E}_{\state \sim p(\state|\action, \theta), \theta \sim \modelpost, \action \sim \mathcal{U}}\left[\action\cdot p(\optimal|\traj)\right]}
 % ------------------------------------------------------------------
 {\mathbb{E}_{\state \sim p(\state|\action, \theta), \theta \sim \modelpost, \action \sim \mathcal{U}}\left[p(\optimal|\traj)\right]}
=
\frac
 {\mathbb{E}_{\action \sim \mathcal{U}}\left[\action \cdot \mathcal{W}(\action)\right]} %
 % -----------------------------------------------------------------------------------
 {\mathbb{E}_{\action \sim \mathcal{U}}\left[\mathcal{W}(\action)\right]}, \label{eqn:mm}
\end{equation}
where
\begin{equation}
    \mathcal{W}(\action) :=  \mathbb{E}_{\state_{t+1} \sim p(\state_{t+1}|\state_{t}, \action_{t}, \theta), \theta \sim \modelpost}\left[p(\optimal|\traj)\right].
\end{equation}
Eq.~\eqref{eqn:mm} can be viewed as a weighted average where each sampled action is weighted by the likelihood of optimality $\mathcal{W}(\action)$.
In the same way, we can also estimate the variance of the posterior
$
\actions = {\mathbb{E}_{\action \sim \mathcal{U}}\left[(\action - \actionm)^{2} \mathcal{W}(\action)\right]}/ %
 % -----------------------------------------------------------------------------------
 {\mathbb{E}_{\action \sim \mathcal{U}}\left[\mathcal{W}(\action)\right]}
$.

In practice, sampling from uniform distribution $\mathcal{U}$ is quite inefficient and requires almost infinite samples.
Hence, let us consider iteratively estimating the parameters by incorporating importance sampling.
Let $\actionm^{(j)}$, $\actions^{(j)}$ be the estimated parameters at iteration $j$; we can rearrange \eqref{eqn:mm} as
\begin{equation}
    \actionm^{(j+1)} \leftarrow
    \{\mathrm{RHS\ of\ \eqref{eqn:mm}}\}
     \times \frac{q(\action; \actionm^{(j)}, \actions^{(j)})}{q(\action; \actionm^{(j)}, \actions^{(j)})}
    =
    \frac
    {\mathbb{E}_{\action \sim q(\action; \actionm^{(j)}, \actions^{(j)})}\left[ \action \cdot \mathcal{W}(\action)\right]} % nominator
    {\mathbb{E}_{\action \sim q(\action; \actionm^{(j)}, \actions^{(j)})}\left[ \mathcal{W}(\action)\right] }. \label{eqn:mm_iter} % denominator
\end{equation}

It is worth noting that a similar iterative law can also be derived by solving the optimization problem
%
% $\operatorname{argmin}_{q} \mathbb{E}_{\traj \sim q(\traj), \theta \sim \modelpost}\left[\log p(\optimal|\traj)\right]$
$\operatorname{argmax}_{q(\action;\actionm, \actions)} \mathbb{E}\left[\log p(\optimal|\traj)\right]$
by mirror descent~\citep{miyashita2018mirror,okada2018acceleration}.
To connect this inference problem to trajectory optimization, we define the optimality likelihood with trajectory reward $r(\traj)$ and a monotonically increasing function $f(\cdot)$, as $p(\optimal|\traj) := f(r(\traj))$.
If we define $f(r(\traj)) \propto e^{r(\traj)}$ the same as~\citep{levine2018reinforcement,piche2018probabilistic}, an optimization algorithm similar to MPPI~\citep{williams2016aggressive,williams2017information,okada2018acceleration} is recovered.
As summarized in \tabref{tab:opt_algs}, other similarities to well-known optimization algorithms, including CEM, can be observed with different optimality definitions. %
\footnote{
We implicitly assume the existence of step-wise likelihood $p(\mathcal{O}_{t}| \state_{t}, \action_{t})$ corresponding to each definition.
Since another graphical model with a single unified optimality can be defined, the existence is not critical.}
\begin{table}[tb]
    \centering
    \caption{Optimization algorithms derived by moment matching of $p(\traj|\mathcal{O})$ and different $f$ definitions; $\mathbbm{1}$ indicates an indicator function, $g: \mathbb{R} \mapsto \mathbb{N}$ denotes rank-preserving transformation.}
    \begin{tabular}{c||c|c|c|c}
         & MPPI~\citep{williams2016aggressive}
         & CEM~\citep{botev2013cross}
         & Prop-CEM~\citep{goschin2013cross}
         & CMA-ES~\citep{hansen2003reducing}
         \\\hline
        $ f(\totalreward)$
        & $\propto e^{\totalreward}$ & $\mathbbm{1}\left[\totalreward > r_{thd}\right]$
        & $\displaystyle \frac{\totalreward - r_{min}}{r_{max}-r_{min}}$
        & $\propto \log g(\totalreward) \cdot \mathbbm{1}\left[\totalreward > r_{thd}\right]$
    \end{tabular}
    \label{tab:opt_algs}
\end{table}

There is a discrepancy between \eqref{eqn:mm_iter} and the CEM implementation in~\citep{chua2018deep};
in which $\mathcal{W'}(\action) = f(\mathbb{E}[r(\traj)])$ is used instead of $\mathcal{W}(\action) = \mathbb{E}[f(r(\traj))]$.
% Let us consider to heuristically replace  to , where
Since $f$ is a convex function, Jensen's inequality holds in this case, thus $\mathcal{W} \geq \mathcal{W}'$.
The equality holds when $f(\cdot)$ is constant, implying that
$\mathcal{W} \simeq \mathcal{W}'$ for low-variance $r(\traj)$ and
$\mathcal{W} > \mathcal{W}'$ for high-variance (or more uncertain) $r(\traj)$.
Namely, $\mathcal{W}'(\action)$ underestimates the optimality likelihood if $\action$ generates uncertain trajectories.
% In vice versa, $\mathcal{W}'(\action)$ exaggerates the likelihood of less uncertain actions.
Since we have experimentally observed that this uncertainty avoidance behavior by $\mathcal{W}'$ demonstrates higher optimization performance than $\mathcal{W}$ (see \secref{sec:comp_w}),
this paper heuristically employs the use of $\mathcal{W}'$.

In practice, expectation operators $\mathbb{E}[\cdot]$ should be implemented on digital computers
through the Monte Carlo integration with $K$ sampled actions and $P$ trajectories for each action:
$\actionm^{(j+1)} \simeq {\sum^{K}_{k=1}\left[ \action_{k} \cdot \mathcal{W'}(\action_{k})\right]}/{\sum^{K}_{k=1}\left[ \mathcal{W'}(\action_{k})\right] }$ and
$\mathcal{W'}(\action_{k}) \simeq f\left(\frac{1}{P}\sum^{P}_{i=1} r(\traj_{k,i})\right)$.

\section{Variational Inference MPC: From Moment Matching to Inference} \label{sec:from_mm_to_vi}
Given uncertainty in a dynamics model, it is natural to suppose that
optimal trajectories are also uncertain.
However, as exhibited in the previous section, PETS employs the moment matching of the trajectory posterior, ignoring almost uncertainty in optimal trajectories.
In this section, we newly introduce a variational inference MPC (VI-MPC) framework to formulate MBRL as fully Bayesian and involve uncertainty both in the dynamics and optimalities.

Let us consider a variational inference problem:
$\operatorname{KL}\left(q_{\theta}(\traj)||p(\traj, \theta|\optimal)\right)$.
We assume the variational distribution $q_{\theta}(\traj)$ is decomposed to
$q_{\theta}(\traj) = q(\action) p(\state|\action, \theta)\modelpost$;
hence, we introduce $p(\traj, \theta|\optimal)(=p(\optimal|\traj)p(\state|\action,\theta)\modelpost)$ as a posterior, which takes the similar decomposable form as $q_{\theta}(\traj)$.
This assumption forces optimal state transitions to be controlled only by $p(\state_{t+1}|\state_{t}, \action_{t}, \theta)$~\citep{levine2018reinforcement}.
% After the variational inference, we marginalize $(\state, \theta)$ from $q_{\theta}(\traj)$.
% Let $q_{\theta}(\traj)$ be a variational distribution to approximate the posterior $p(\traj, \theta|\optimal)$.
% The objective of variational inference here is to find $q_{\theta}(\traj)$ that minimizes
%
% $\operatorname{argmin}_{q}\operatorname{KL}\left(q(\traj)||p(\traj|\optimal)\right)$.
% $\operatorname{KL}\left(q_{\theta}(\traj)||p(\traj, \theta|\optimal)\right)$.
%
% Assuming that $q_{\theta}(\traj)$ can be decomposed into $q_{\theta}(\traj) = q(\action) p(\state|\action, \theta)\modelpost$,
As shown in \secref{sec:derivation_elbo}, this inference problem can be transformed to the maximization problem:
$\operatorname{argmax}_{q_{\theta}(\traj)} \mathbb{E}\left[\log p(\optimal|\traj) - \log q(\action) \right]$.
A notable property is that this objective has an entropy regularization term $-\log q(\action)$, which encourages $q(\action)$ to have broader shape to capture more uncertainty.
For the sake of convenience, we introduce a tunable hyperparameter $\alpha (> 0)$ to the optimality likelihood $p(\optimal|\traj) \rightarrow p^{\frac{1}{\alpha}}(\optimal|\traj)$.
Then the above objective can be transformed as
$\operatorname{argmax}_{q_{\theta}(\traj)} \mathbb{E}\left[\log p(\optimal|\traj) - \alpha \log q(\action) \right]$.
By applying mirror descent~\citep{bubeck2015convex} to this optimization problem, we can derive an update law for $q(\action)$ (see \secref{sec:derivation_eq5} for the detailed derivation):
\begin{equation}
    q^{(j+1)}(\action) \leftarrow
    % \frac
    {q^{(j)}(\action)\cdot \mathcal{W}'(\action)^{\frac{1}{\lambda}}\cdot (q^{(j)}(\action))^{-\kappa}}
    % =================================================================================================
    {\Big /}
    {\mathbb{E}_{q^{(j)}(\action)}\left[\mathcal{W}'(\action)^{\frac{1}{\lambda}}\cdot (q^{(j)}(\action))^{-\kappa}\right]}, \label{eqn:vi}
\end{equation}
where $\lambda (>0)$, $\kappa (>0)$ are hyperparameters and $\alpha$ is absorbed into them. $\lambda$ is inverted step-size to control optimization speed and $\kappa$ is the weight of the entropy regularization term $q^{-\kappa}$.

Eq.~(\ref{eqn:vi}) suggests a novel and general MPC framework, which we call variational inference MPC ({VI-MPC}).
To realize a specific VI-MPC method, we specify the following parameters:
(1) optimality definition (or $f(\cdot)$; see \tabref{tab:opt_algs}), (2) variational distribution model $q$, and (3) entropy regularization $\kappa > 0$ or $\kappa = 0$. %
We did not include $\lambda$ into the specifications since it is highly dependent on the optimality definition (see \secref{sec:impl_notes}).
In this paper, we describe the above specifications as
\texttt{VIMPC({<optimality\_def>}, {<variational\_dist>}, <max\_ent>)}.
For example, we respectively express vanilla CEM and MPPI as
\texttt{VIMPC(`CEM', `Gaussian', False)}
and
\texttt{VIMPC(`MPPI', `Gaussian', False)}.
In \secref{sec:vimpc}, we propose a new instance of VI-MPC to incorporate multimodal uncertainty in the posterior. % $p(\traj|\mathcal{O})$.
%}
% In~\citep{chua2018deep}, $K=500$, $P=20$ is used. We also use these values.

% \begin{quote}
% In this section, we show how SMCP can deal with multimodal policies when planning. We believe
% multimodality is useful for exploring since it allows us to keep a distribution over many promising
% trajectories and also allows us to adapt to changes in the environment e.g. if a path is suddenly
% blocked.
% \end{quote}

%===============================================================================
\section{Probabilistic Action Ensembles with Trajectory Sampling} \label{sec:vimpc}
As reviewed in \secref{sec:dynamics_inference}, previous methods have successfully involved multimodality in $\modelpost$ with network ensembles.
If this multimodality in $\modelpost$ is given, other distributions depending on $\modelpost$, including $p(\optimal|\traj)$, would also be multimodal.
In other words, there are various possible optimal trajectories (or actions) like \figref{fig:toytask}.
It is obvious that \texttt{VIMPC(*, `Gaussian', *)} will still easily fail to capture multimodality because of overfitting to a single mode.
Inspired by the success of the ensemble approach for dynamics modeling, we propose a novel VI-MPC method that introduces action ensembles with a Gaussian mixture model (GMM), i.e.,~\texttt{VIMPC(*, `GMM(M=*)', *)}, which
we call \proposedmethod{} (Probabilistic Action Ensembles with Trajectory Sampling).

\proposedmethod{} defines the variational distribution $q(\action)$ as
\begin{equation}
q^{(j)}(\action) := q(\action; \phi^{(j)}) = \sum^{M}_{m=1}\pi^{(j)}_{m} \mathcal{N}(\action; \actionm^{(j)}_{m}, \actions^{(j)}_{m}), \label{eqn:q_gmm}
\end{equation}
where
$\phi^{(j)} := \{(\mixcoef_{m}^{(j)}, \actionm_{m}^{(j)}, \actions_{m}^{(j)} )\}^{M}_{m=1}$ and $M$ is the number of components of the mixture model.
Now, we derive the iteration scheme to update the parameters of GMM.
At first, drawing $K$ samples from $q^{(j)}(\action)$, we approximate $q^{(j)}(\action)$ as a discretized distribution (or a set of particles):
\begin{equation}
q^{(j)}(\action; \phi) \simeq q(\action; \mathbf{W}^{(j)}) := \sum^{K}_{k=1}w_{k}^{(j)} \delta(\action - \action_{k}),
\end{equation}
where $\mathbf{W}^{(j)} := \{w_{k}^{(j)} \}_{k=1}^{K}$.
Just after sampling, the weight of each particle is uniform:
$\mathbf{W}^{(j)} = \mathbf{1}/K$.
By substituting this approximated distribution to \eqref{eqn:vi}, the update law for the particle weights is derived as
\begin{equation}
    w_{k}^{(j+1)} \leftarrow
    % \frac
    {\mathcal{W}'(\action_{k})^{\frac{1}{\lambda}}\cdot (q^{(j)}(\action_{k}))^{-\kappa}}
    % ==================================================================================================
    {\Big /}
    {\sum_{k'=1}^{K}\mathcal{W}'(\action_{k'})^{\frac{1}{\lambda}}\cdot (q^{(j)}(\action_{k'}))^{-\kappa}}. \label{eqn:weight_update}
\end{equation}
%
% Likelihood that we observe this weighted particles
Then we estimate $\phi^{(j+1)}$, which maximizes the observation probability of the weighted particles:
\begin{equation}
\phi^{(j+1)} = \operatorname{argmax}_{\phi} \log p(\{(w^{(j+1)}_{k}, \action_{k})\}_{k=1}^{K} |\phi) =
\operatorname{argmax}_{\phi}
\sum^{K}_{k=1}
w_{k}^{(j+1)}
\log q(\action_{k}; \phi).
% \operatorname{argmax}_{\phi}
% \sum^{K}_{k=1}
% \log \left(
% w_{k}^{(j+1)}
% \sum^{M}_{m=1}\pi_{m} \mathcal{N}(\action_{k}; \actionm_{m}, \actions_{m})
% \right).
\end{equation}
% \{(w^{(j+1)}_{k}, \action_{k})\}_{k=1}^{K}|\phi
By taking the derivative $\nabla_{\phi}\log p(\cdot|\phi) = \mathbf{0}$ and borrowing the concept of the EM algorithm~\citep{bilmes1998gentle}, we get the update laws of $\phi^{(j+1)}$ which take the weight-average form like \eqref{eqn:mm_iter} (see \secref{sec:paets_eqs} for the complete definition):
\begin{equation}
% \begin{align}
\left(\actionm_{m}^{(j+1)}, \actions_{m}^{(j+1)}, \mixcoef_{m}^{(j+1)} \right)
\leftarrow
\left(
\sum^{K}_{k=1} \omega^{(j+1)}_{m,k}\action_{k},
{\sum^{K}_{k=1} \omega^{(j+1)}_{m,k}(\action_{k} - \actionm_{m}^{(j+1)})^{2}},
\frac{N_{m}}{\sum^{M}_{m=1} N_{m}}
% {N_{m}}/{\sum^{M}_{m=1} N_{m}}
\right). \label{eqn:paets}
    % \actionm_{m}^{(j+1)} \leftarrow
    %     \sum^{K}_{k=1} \omega^{(j+1)}_{m,k}\action_{k},
%
    % \actions_{m}^{(j+1)} &\leftarrow&
    %     {\sum^{K}_{k=1} \omega^{(j+1)}_{m,k}(\action_{k} - \actionm_{m}^{(j+1)})(\action_{k} - \actionm_{m}^{(j+1)})^{\mathrm{T}}} \\
    % \actions_{m}^{(j+1)} &\leftarrow
    %     \sum^{K}_{k=1} \omega^{(j+1)}_{m,k}(\action_{k} - \actionm_{m}^{(j+1)})(\action_{k} - \actionm_{m}^{(j+1)})^{\mathrm{T}} \\
%
    % \mixcoef_{m}^{(j+1)} \leftarrow N_{m} {\Big /} \sum^{M}_{m=1} N_{m}. \label{eqn:mix_coef}
\end{equation}
\begin{wrapfigure}{r}{0.4\textwidth}
    \centering
    \includegraphics[width=0.35\textwidth]{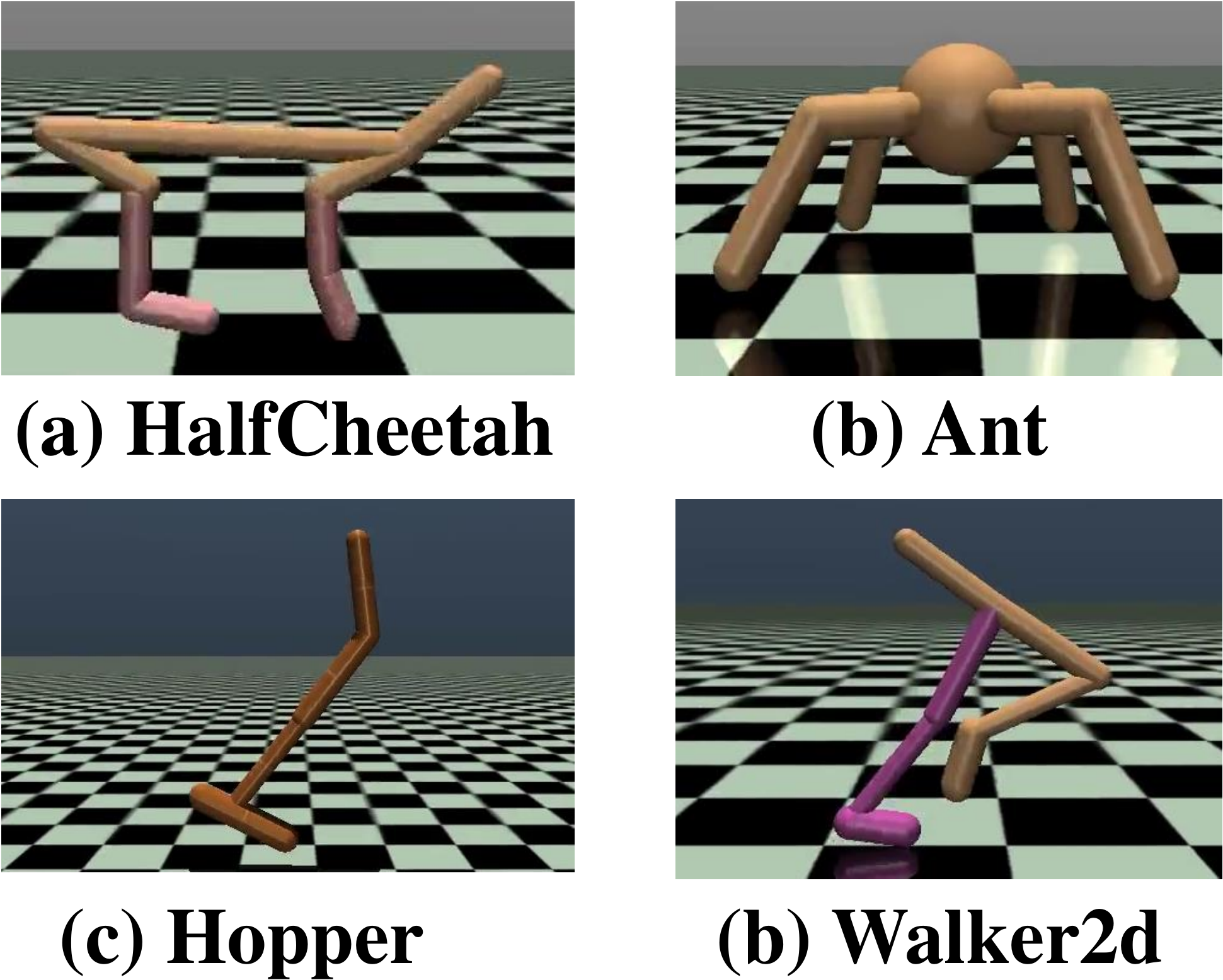}
    \vspace*{-2.5mm}
    \caption{Evaluated locomotion tasks simulated in MuJoCo.} \label{fig:mujoco}
\end{wrapfigure}
\figref{fig:toytask_opt} in \secref{sec:toytask_opt} illustrates how this method works in a toy optimization task.

In summary, \proposedmethod{} and MBRL utilizing it are respectively described in Algs.~\ref{alg:vi_traj_opt} and \ref{alg:vi_mpc}, where $U$ is the number of iterations for optimization and $H$ is the length of the task episode.
% Alg.~\ref{alg:vi_mpc} represents a single iteration of \textit{training-} and \textit{test-step}, and is iteratively executed.
% Before executing the first iteration, $\mathcal{D}$ is initialized to be $|\mathcal{D}|=H$ with a random controller.
At $\ell 4$ in Alg.~\ref{alg:vi_mpc}, $\actionm_{m}$s are initialized independently at random.
% We experimentally observed that this is sufficient to involve the multimodality.
% At $\ell 10$ in Alg.~\ref{alg:vi_mpc}, general technique in MPC, so called \textit{warm startup}, is introduced to re-initialize $\actionm_{m}$s.
% At the following line $\ell 11$, $\actions_{m}$s and $\mixcoef_{m}$s are reset to be initial values,
At $\ell 12$, $\actions_{m}$s and $\mixcoef_{m}$s are reset to be initial values,
encouraging exploration for the next time-step and preventing $q(\action; \phi)$ from degenerating to a single mode.
If we set $M=1$, these procedures are almost equivalent to those of PETS.
The use of GMM ($M>1$) does not increase computational complexity significantly (see \secref{sec:comp_comp}).
% The discussions about the computational complexity of our method is found in .
%
% \begin{wrapfigure}{r}{0.5\textwidth}
\newcommand\mycommfont[1]{\scriptsize\ttfamily{\color[gray]{0.5}{#1}}}
\SetCommentSty{mycommfont}
\SetKwRepeat{Do}{do}{while}%

\begin{figure}[tb]
\begin{minipage}[t]{0.46\textwidth}
\begin{algorithm}[H]
\footnotesize
\DontPrintSemicolon
  \KwInput{State $\state_{1}$, GMM param.~$\phi^{(1)}$ and $\modelpost$}
  \KwOutput{Optimized GMM param.~$\phi^{(U+1)}$}
  \For{$j \leftarrow 1$ \KwTo $U$}{
%   Sample actions $\{\action_{k} \sim q(\action; \phi^{(j)})\}_{k=1}^{K}$ \;
%   Sample trajectories $\{\traj_{k, i} \sim p(\state; \action, \theta)\modelpost \}_{k=1}^{K}$ \;
  Sample actions $\{\action_{k} \sim q(\action; \phi^{(j)})\}_{k=1}^{K}$ \;
  Sample states $\{ \{ \{$ \;
  \Indp $\theta_{k,i,t} \sim \modelpost $ \tcp*[l]{TS1 method}
  $\state_{k, i,t+1} \sim p(\state_{t+1}|\action_{k,t}, \state_{k, i, t}, \theta_{k,i,t})$, \;
  \Indm $\}_{t=1}^{T-1} \}_{i=1}^{P}  \}_{k=1}^{K}$ \;
    %  \
    % $\{\state_{k,i} \sim p(\state | \action_{k}, \theta)\modelpost \}_{i=1}^{P}$ \;
  Eval.~$\{\mathcal{W}'(\action_{k}) \simeq f(\sum^{P}_{i=1}r(\traj_{k, i}))\}_{k=1}^{K}$ \;
  Calc.~$\{w_{k}^{(j+1)}\}^{K}_{k=1}$ by \eqref{eqn:weight_update} \;
  Update $\phi^{(j+1)}$ by \eqref{eqn:paets}
  }
\caption{\proposedmethod} \label{alg:vi_traj_opt}
\end{algorithm}
\end{minipage}
\hspace{0.04\textwidth}
% \end{wrapfigure}
% \begin{wrapfigure}{r}{0.5\textwidth}
\begin{minipage}[t]{0.5\textwidth}
\begin{algorithm}[H]
\footnotesize
\DontPrintSemicolon
  % \KwInput{Initial state $\state_{1}$}
  \KwData{initial variance ~$\actions_{init}$}
  Init.~$\mathcal{D}$ with a random controller for one trial \;
  \Repeat{the MPC-policy performs well}{
  Infer $p_{\mathcal{D}}(\theta)$ \tcp*[l]{train ensemble DNNs}
%   \For{$m \leftarrow 1$ \KwTo $M$ \tcp*{initialization of $\phi$}}{
    $\{\actionm_{m} \leftarrow \mathcal{N}(\action; \mathbf{0}, \actions_{init})\}^{M}_{m=1}$ \tcp*[l]{rand.~init.}
    $\{(\actions_{m}, \mixcoef_{m}) \leftarrow (\actions_{init}, 1/M)\}^{M}_{m=1}$ \;
%   }
  \For{$n \leftarrow 1$ \KwTo $H$}{
  $\phi \leftarrow $Exec.~Alg.~\ref{alg:vi_traj_opt}$(\state_{n}, \phi, \modelpost)$\;
  Sample $\action \sim q(\action; \phi)$ \;
  Send $\action_{1}$ to actuators and observe $\state_{n+1}$\;
  $\mathcal{D} \leftarrow \mathcal{D} \cup \{(\state_{n}, \action_{1}, \state_{n+1})\}$ \;
%   \For{$m \leftarrow 1$ \KwTo $M$ \tcp*{re-init.~of $\phi$}}{
  $\{\actionm_{m} \leftarrow \{\actionm_{m, 2:T}, \mathbf{0}\}\}_{m=1}^{M}$ \tcp*[l]{warm-start}
  $\{(\actions_{m}, \mixcoef_{m}) \leftarrow (\actions_{init}, 1/M)\}^{M}_{m=1}$ \;
    % $\actions_{m} \leftarrow \actions_{init}$ \;
    % $\mixcoef_{m} \leftarrow 1/M$ \;
%   }
%   \If{Task completed}{
%   Break this loop \;
%   }
  }}
\caption{MBRL with \proposedmethod} \label{alg:vi_mpc}
\end{algorithm}
\end{minipage}
% \end{wrapfigure}
\end{figure}

\section{Experiments} \label{sec:experiment}
\subsection{Comparison to State-of-the-art Methods}
\begin{figure}[t]
    \centering
    \includegraphics[width=0.9\textwidth]{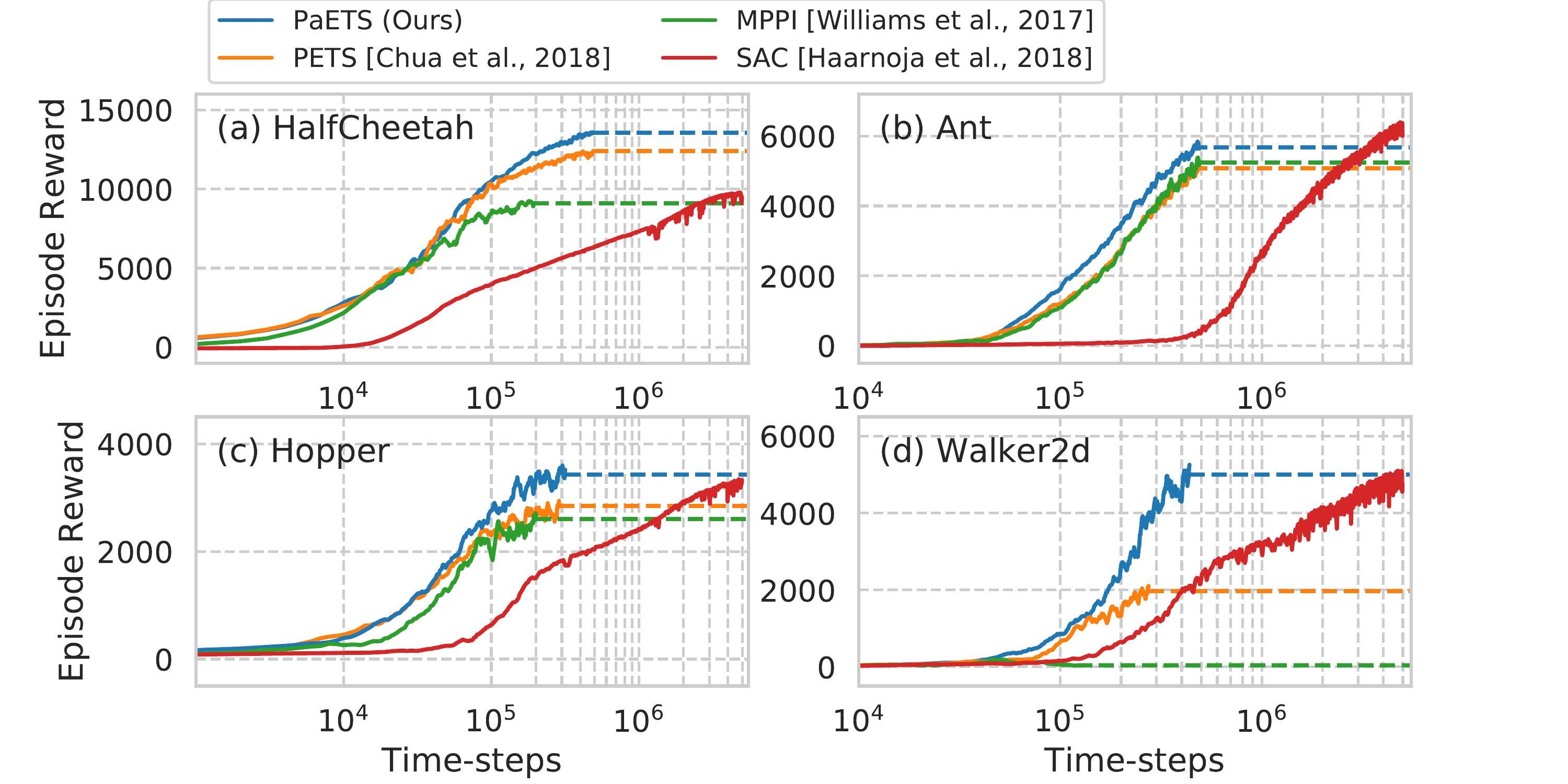}
    \vspace{-2.5mm}
    \caption{Learning curves for different tasks and algorithms.
    These are averaged results of 8 (for MBRL) and 20 (for SAC) trials with different random seeds.
    We stopped the training when convergence was observed or after reaching the specified test steps ($500$ for MBRL and $5,000$ for SAC).
    The asymptotic performances (averages of the last 10 test steps) are depicted in dashed lines.
    }
    \label{fig:learning_curves}
\end{figure}

The main objective of this experiment is to demonstrate that \proposedmethod{} has advantages over the state-of-the-art MBRL baseline: PETS~\citep{chua2018deep}.
In this experiment, \proposedmethod{} and PETS (or vanilla CEM) were implemented
using our same codebase with different parameters, i.e.,
\texttt{VIMPC(`CEM', `GMM(M=5)', True)} for \proposedmethod{}, and \texttt{VIMPC(`CEM', `GMM(M=1)', False)} for PETS.
We also evaluated another MBRL baseline with MPPI ~\citep{williams2017information}, realized as \texttt{VIMPC(`MPPI', `GMM(M=1)', False}).
These above methods share the settings for $\modelpost$ inference (training of network ensembles).
The state-of-the-art MFRL method SAC~\citep{haarnoja2018soft}, was also evaluated to compare asymptotic performance.%%between MBRL and MFRL.
\footnote
{We used the open-source code: \url{https://github.com/pranz24/pytorch-soft-actor-critic}
}
\figref{fig:mujoco} illustrates the simulated locomotion tasks evaluated in this experiment, which are complex and challenging due to their high non-linearity.
% All the tasks, except for HalfCheetah, were not evaluated in the original PETS paper~\citep{chua2018deep}.
Other details about our implementation and experimental settings are described in \secref{sec:impl_notes} and \secref{sec:exp_setup}.
\figref{fig:learning_curves} presents the experimental results, in which \proposedmethod{} consistently exhibits better asymptotic performance than that of the  MBRL baselines.
In addition, \proposedmethod{} outperforms or is comparable to SAC while requiring significantly fewer samples (about x10 more sample efficient).
% Although other realizations for \proposedmethod{} are possible, e.g.,~\texttt{VIMPC(`MPPI', `GMM(M=5)', True)}, we did not include their results since CEM-based realization consistently showed the best performances.
%

\subsection{Ablation Study}
\begin{figure}[t]
    \centering
    \includegraphics[width=0.9\textwidth]{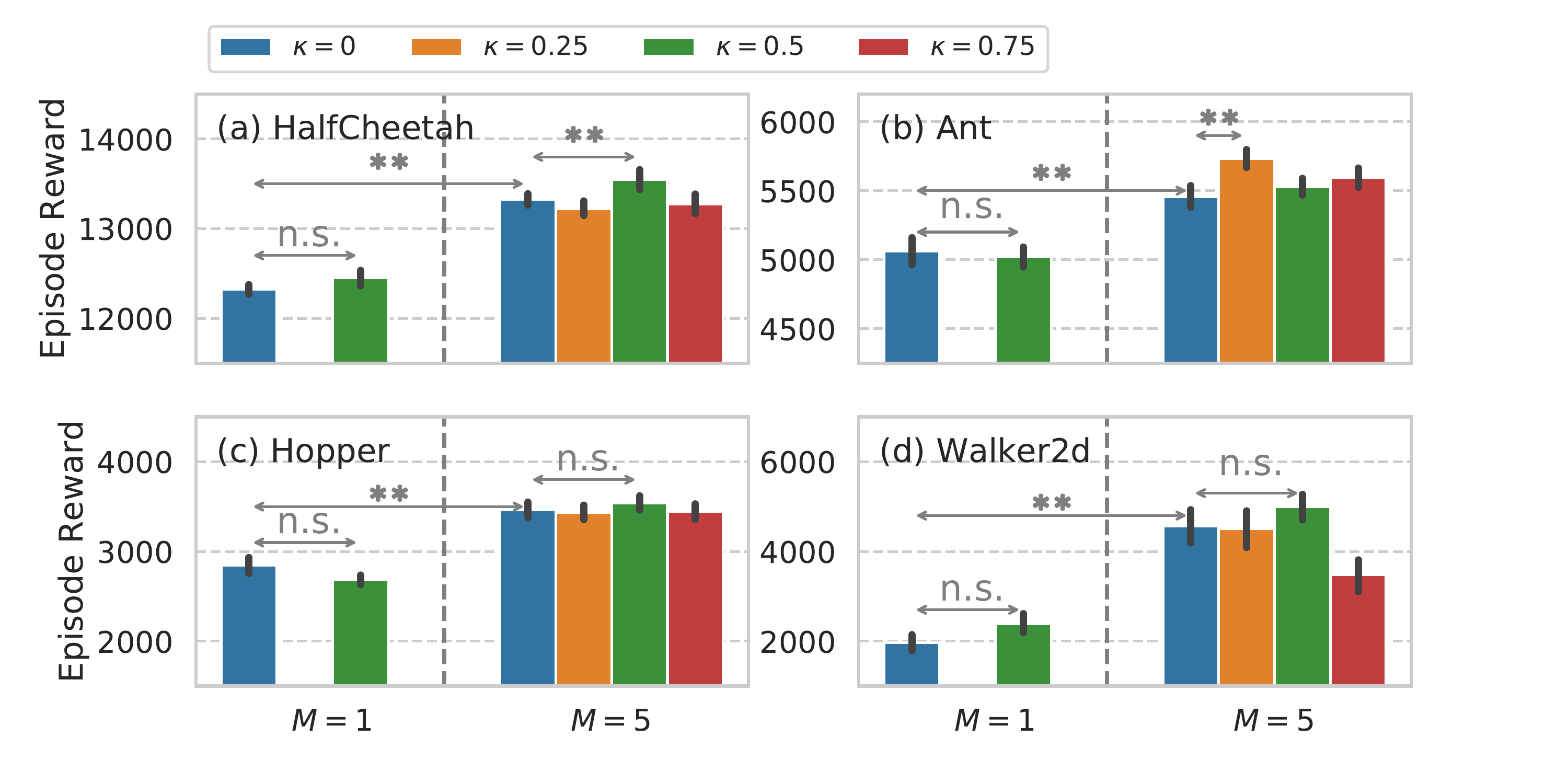}
    \vspace*{-5mm}
    \caption{
    Asymptotic performance comparison with varying $M$s and $\kappa$s.
    These are averaged results over 8 different MBRL trials and the last 10 test steps.
    The error bars denote confidence intervals (95\%).
    Symbols `**' and `n.s.' respectively mean $p<0.01$ and $p \geq 0.01$ in Welch's $t$-test.
    }
    \label{fig:kappa_analysis}
\end{figure}

\begin{wrapfigure}{r}{0.45\textwidth}
    \centering
    \includegraphics[width=0.4\textwidth]{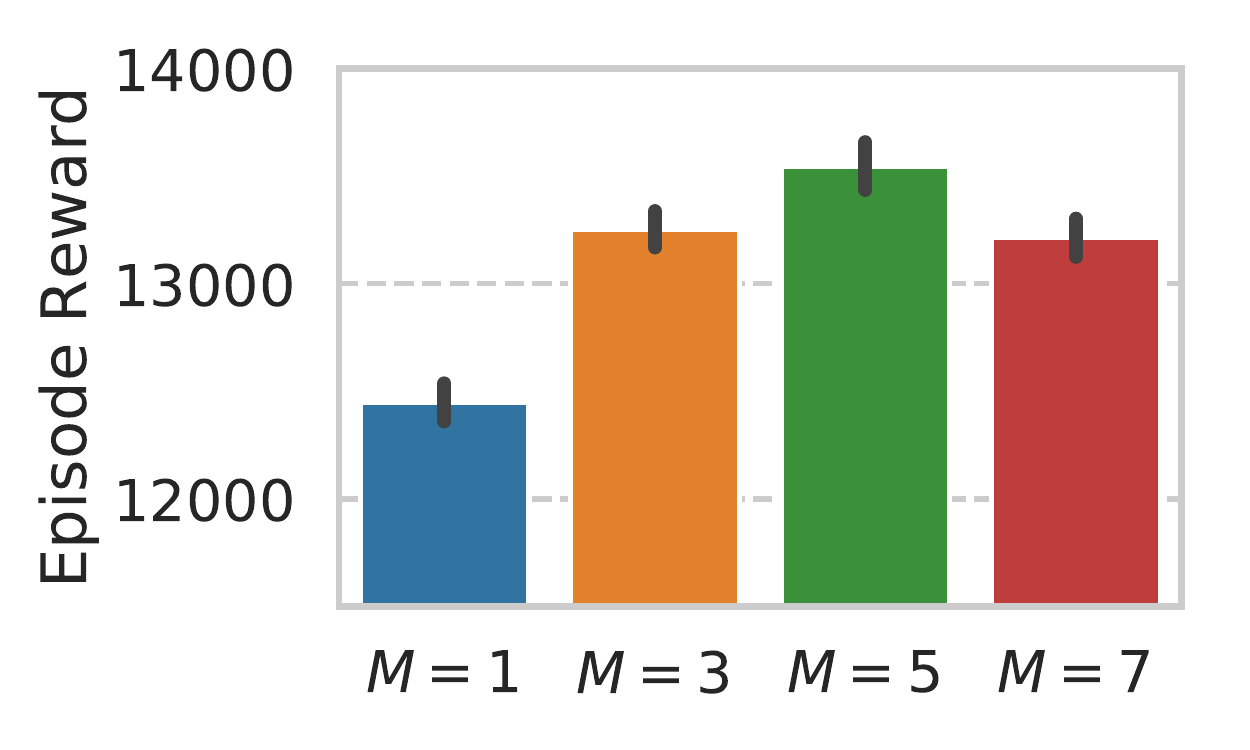}
    \vspace{-4mm}
    \caption{
    Asymptotic performance comparison with varying $M$s. % and fixed $\kappa(=0.5)$.
    Only the HalfCheetah task is evaluated in this test.
    % Significant improvement ($p<0.001$) by GMM ($M > 1$) is observed for all cases.
    }
    \label{fig:gmm_size_analysis}
\end{wrapfigure}

This experiment clarifies which component of \proposedmethod{} (GMM and entropy-regularization) contributed to the overall improvement.
\figref{fig:kappa_analysis} expresses the results of this ablation study and Welch's $t$-test for some selected representative pairs.
From this figure, one can observe that the use of GMM ($M=5$) significantly improves performance.
The effect of the regularization ($\kappa > 0$) is relatively small, but not negligible.
In certain tasks, setting $\kappa$ to particular values could improve the performance.
% In the case of $M=1$, the entropy regularization effects only the magnitude of variace.
In the case of $M>1$, the regularization sheds light on actions sampled from low $\mixcoef_{m}$, thus encouraging $q(\action; \phi)$ to be multimodal.
In some tasks which requires rather delicate controls (e.g., Hopper, Walker2d), the effect of $\kappa$ seems less significant.
\figref{fig:gmm_size_analysis} examines sensitivity with the number of mixture components $M$, for which $M=5$ achieved the highest performance.
If infinite or enough samples are given ($K \gg 0$),
it would be reasonable to set $M$ to be large enough to capture multimodality.
However, in practice, $K$ is finite and could be small enough due to computational constraints.
In this case, larger $M$ makes it difficult to approximate $q(\action; \phi)$ as a set of particles $q(\action; \mathbf{W})$, resulting in degradation of the optimization performance.


\section{Related Work}
%
%
% Kernel density estimation~\citep{chen2017tutorial}
\textbf{Dynamics Posterior Inference}
Recent MBRL methods, MB-MPO (Model-Based Meta-Policy-Optimization)~\citep{clavera2018model} and ME-TRPO (Model Ensemble Trust Region Optimization)~\citep{kurutach2018model}, also employ network ensembles to model dynamics, but they utilize the ensembles differently than we do: to train policy networks, not MPC.

\textbf{Trajectory Optimization}
Sequential Monte-Carlo based MPC, described as \texttt{VIMPC(*, `Particles', False)}, has been introduced in \citep{kantas2009sequential}, but it requires well-designed proposal distribution to sample particles for the next iteration~$j+1$.
% In addition, evaluation is limited to toy tasks.
Another particle-based method has been derived~\citep{piche2018probabilistic}
by utilizing the \textit{control as inference} framework.
However, this method relies on not only a dynamics model, but also policy and value functions to manage particles, so MFRL methods must be incorporated.

Recent studies have demonstrated that entropy regularization is a promising strategy in policy training~ \citep{abdolmaleki2015model,abdolmaleki2017deriving,haarnoja2017reinforcement,haarnoja2018soft}.
% Particularly, SAC~\citep{haarnoja2018soft} has been regarded as one of the most powerful baselines in recent MFRL fields.
However, to the best of our knowledge, the introduction of entropy regularization to MPC is novel along with explicit multimodal expression to successfully realize their synergistic effect.

Ref.~\citep{wagener2019online} also systematically organizes the stochastic MPC methods from the perspective of online learning, but uncertainty-aware discussions from a Bayesian viewpoint are not conducted.

\textbf{Bayesian Reformulation}
Ref.~\citep{jeon2018bayesian} proposes a novel approach to generative adversarial imitation learning (GAIL)~\citep{ho2016generative}, which reformulates general GAIL in a Bayesian fashion and utilizes ensembles to infer discriminator posteriors. %significantly improving the sample efficiency.
Another Bayesian reformulation of GAIL integrates imitation and reinforcement learning by introducing another optimality (i.e.,~imitation optimality $\mathcal{O}_{t}^{I}$)~\citep{kinose2019integration}.

% \section{Discussions}
\section{Conclusion \& Discussions}
This paper introduces a novel VI-MPC framework that systematically generalizes and reformulates various stochastic MPC methods in a Bayesian fashion.
We also devise a novel instance of this framework, called \proposedmethod{}, which can successfully incorporate multimodal uncertainty in optimal trajectories.
By combining our method and the recent uncertainty-aware dynamics modeling with neural network ensembles,
our Bayesian MBRL is able to involve multimodalities both in dynamics and optimalities.
In addition, our method is a quite simple extension of general stochastic methods and requires no significant additional computational complexity.
Our experiments demonstrate that \proposedmethod{} can improve asymptotic performance compared to the leading MBRL baseline PETS,
and thus substantially enhances MBRL potential to be more competitive to the state-of-the-art MFRL.
% From these results, we believe that this work significantly contributed to improve the possibilities of MBRL.
% In similar to other recent MBRF methods, our method is also sample efficient than MFRL.

Considering the simplicity and generalizability of VI-MPC and \proposedmethod{},
we expect that our concept is applicable to a variety of tasks,
such as traditional MPC with deterministic dynamics and advanced MPC with latent dynamics from pixels by Deep Planning Network ~\citep{hafner2018learning}.
By introducing a categorical mixture model as a variational distribution, application to combinational optimizations is also feasible.
In fact, our ongoing work includes experiments of discrete MPC for a practical system.

A question that remains is how to determine VI-MPC specifications.
As implied in \figref{fig:learning_curves}, the best optimality definition could be task dependent
(e.g., MPPI outperformed vanilla CEM in the Ant but not in other tasks).
The regularization weight $\kappa$ also has task dependency as shown in \figref{fig:kappa_analysis}.
It would be challenging but interesting future work to add the parameters to the graphical model in \figref{fig:graphical_model} as latent variables
to infer promising parameters along with optimal trajectories, like infinite GMM~\citep{rasmussen2000infinite}.
Another appealing endeavor for future work is to introduce the concept of parallel tempering~\citep{brooks2011handbook} in Markov Chain Monte Carlo.
By adaptively varying different temperatures ($\lambda$ in our case) of ensemble actions,
we can expect the ensemble diversity to improve.
%===============================================================================

\clearpage

% The acknowledgments are automatically included only in the final version of the paper.
\acknowledgments{
We thank Vishwajeet Singh, Hiroki Nakamura and Akira Kinose for their cooperation in this study during their student-internship periods.
Most of the experiments were conducted in ABCI (\textit{AI Bridging Cloud Infrastructure}), built by the National Institute of Advanced Industrial Science and Technology, Japan.
% If a paper is accepted, the final camera-ready version will (and probably should) include acknowledgments. All acknowledgments go at the end of the paper, including thanks to reviewers who gave useful comments, to colleagues who contributed to the ideas, and to funding agencies and corporate sponsors that provided financial support.
}

\bibliography{corl}  % .bib

\appendix

\clearpage

\section{Preliminary Experiment for Uncertainty Modeling} \label{sec:preliminary}
\begin{figure}[h]
  \centering
  \includegraphics[width=0.8\textwidth]{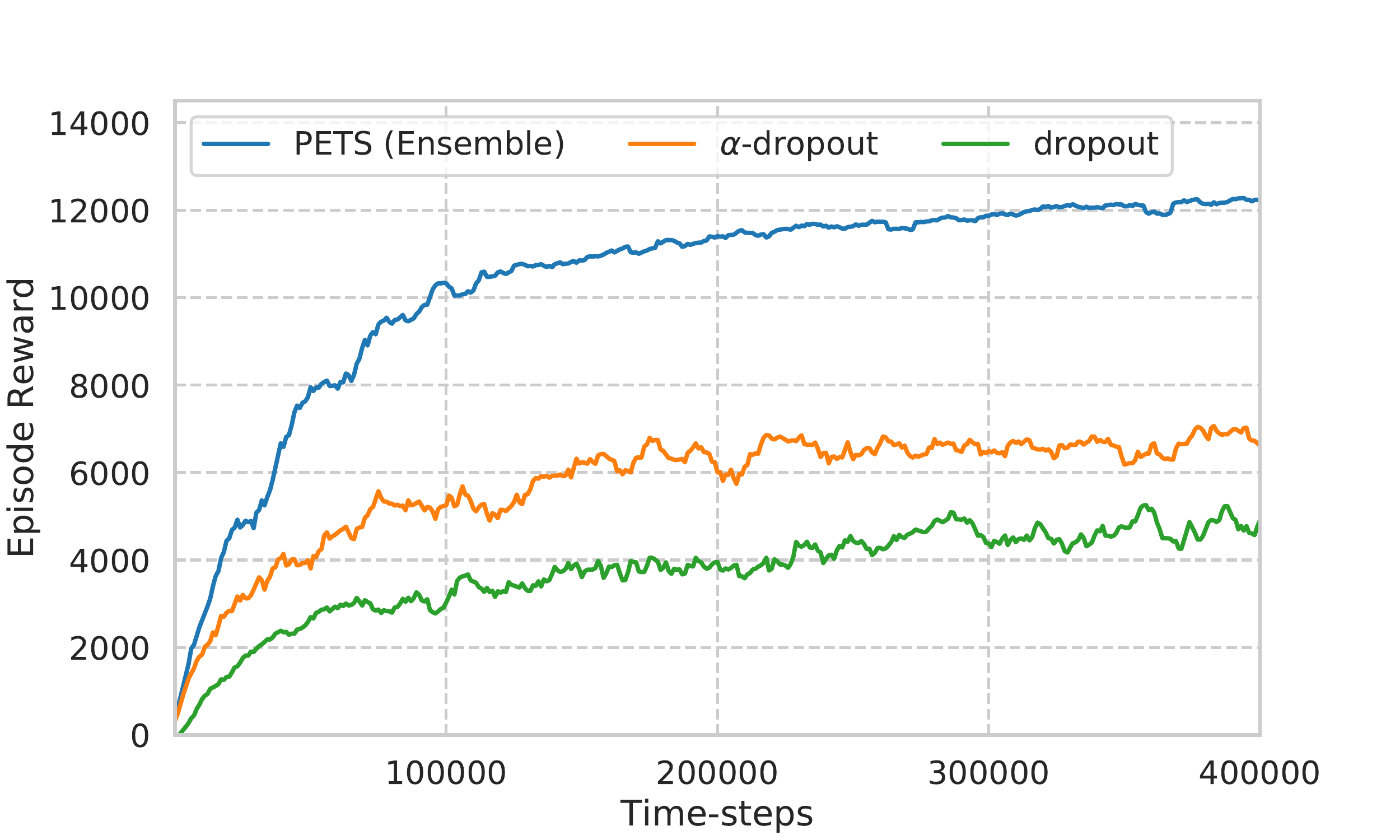}
  \caption{Comparison of uncertainty modeling approaches: ensemble and ($\alpha$-)dropout.}
  \label{fig:preliminary}
\end{figure}

\figref{fig:preliminary} shows the result of a preliminary experiment, in which different uncertainty modeling approaches were evaluated on the HalfCheetah task.
For all trials, vanilla CEM was introduced for trajectory optimization.
This result suggests that ($\alpha$-)dropout is insufficient to capture uncertainty in dynamics, resulting in worse local optima.

\section{Comparison Between $\mathcal{W}$ and $\mathcal{W}'$} \label{sec:comp_w}
We evaluated the impact of $\mathcal{W}$ and $\mathcal{W}'$ on the optimization performance of (vanilla) CEM and MPPI,
the results of which are summarized in \tabref{tab:comp_ws}, where $\mathcal{W}'$ gained much higher rewards than $\mathcal{W}$.
\begin{table}[h]
  \centering
  \caption{
  Episode reward of HalfCheetah task with $\mathcal{W}$ and $\mathcal{W}'$.
  A common dynamics model (sufficiently trained ensemble neural network by MBRL) was employed for this test.
  Ten different trials were conducted and the results were averaged.
  } \label{tab:comp_ws}
  \begin{tabular}{c|c|c|c}
    \multicolumn{2}{c|}{CEM} & \multicolumn{2}{|c}{MPPI} \\\hline
    $\mathcal{W}$ & $\mathcal{W}'$ & $\mathcal{W}$ &  $\mathcal{W}'$ \\\hline
    $5603.24 \pm 541.31$ & $\mathbf{11843.05 \pm 295.80}$  & $2789.03 \pm 647.82$ & $\mathbf{9765.27 \pm 231.04}$
  \end{tabular}
\end{table}

\section{Derivations} \label{sec:derivation}
\subsection{Derivation of the Variational Inference Objective} \label{sec:derivation_elbo}
By using the assumption of $q_{\theta}(\traj) = q(\action)p(\state|\action,\theta)\modelpost$, the KL-divergence can be transformed as
\begin{align}
  \operatorname{KL}\left(q_{\theta}(\traj)||p(\traj, \theta|\optimal)\right)
  &= \int q_{\theta}(\traj) \log \frac{q_{\theta}(\traj)}{p(\traj, \theta | \optimal)} d\traj d\theta \\
  &= \int q_{\theta}(\traj) \log \frac{q(\action)p(\state|\action,\theta)\modelpost}{p(\optimal|\traj)p(\state|\action,\theta)\modelpost} d\traj d\theta \\
  &= - \mathbb{E}_{q_{\theta}(\traj)}\left[\log p(\optimal|\traj) - \log q(\action) \right].
\end{align}

% %
% Similar derivation has been already shown in~\citep{levine2018reinforcement}.
% For clarity purpose, we give the derivation for our MBRL formulation.
% The evidence lower bound (ELBO) is given by
% %
% \begin{align}
%   \log p(\optimal)
%     &= \log \int p(\optimal, \traj, \theta) d\traj d\theta \\
%     &= \log \int p(\optimal, \traj, \theta) \frac{q_{\theta}(\traj)}{q_{\theta}(\traj)} d\traj d\theta \label{eqn:jensen_lhs} \\
%     &\geq \int q_{\theta}(\traj) \log \frac{p(\optimal, \traj, \theta)}{q_{\theta}(\traj)} d\traj d\theta \label{eqn:jensen_rhs} \\
%     &= \int q_{\theta}(\traj) \log \frac{p(\optimal|\traj)p(\state|\action,\theta)\modelpost}{q(\action)p(\state|\action,\theta)\modelpost} d\traj d\theta \\
%     &= \mathbb{E}_{q_{\theta}(\traj)}\left[\log p(\optimal|\traj) - \log q(\action)\right], \label{eqn:elbo}
%     % &\propto \mathbb{E}_{q_{\theta}(\traj)}\left[\log p(\optimal|\traj) - \alpha \log q(\action)\right]
% \end{align}
% %
% where we applied Jensen's inequality in \eqref{eqn:jensen_rhs}.
% By taking the difference between $\log p(\optimal)$ and \eqref{eqn:jensen_rhs},
% we can show that the ELBO maximization is equivalent to
% $\operatorname{argmin}_{q_{\theta}(\traj)} \operatorname{KL}\left(q_{\theta}(\traj)||p(\traj, \theta|\optimal)\right)$.

% \subsection{Derivation of \eqref{eqn:vi}} \label{sec:derivation_eq5}
\subsection{Derivation of \eqref{eqn:vi}} \label{sec:derivation_eq5}
In this section, we simply denote $q_{\action}$ as $q(\action)$ and $q_{\traj}$ as $q(\traj) (= \qa p(\state|\action, \theta)\modelpost)$ for readability.
Let us consider the optimization problem:
\begin{equation}
\operatorname{argmin}_{\qt}\mathcal{J} = \operatorname{argmin}_{\qt} \mathbb{E}_{\qt}\left[-\log p(\optimal|\traj) + \alpha \log \qa \right].
\end{equation}
By applying mirror descent~\citep{bubeck2015convex}, the iterative update law of $\qt^{(j+1)}$ is given as
\begin{equation}
    \qt^{(j+1)} = \operatorname{argmin}_{\qt} \langle \nabla_{\qt}\mathcal{J}, \qt\rangle + \beta\cdot \operatorname{KL}(\qt||\qt^{(j)}) + \gamma \left(1 - \int \qt \cdot d\traj d\theta \right),
\end{equation}
where $\langle \cdot, \cdot \rangle$ is the inner-product operator, $\beta$ is a hyper-parameter related to the step-size, and $\gamma$ is the Lagrange multiplier for the constraint
$\int\qt \cdot d\traj d\theta=1$.
The arguments in the $\operatorname{argmin}$ operator can be rearranged as
\begin{equation}
\int \qt \cdot \left( - \log p(\optimal|\traj) + \alpha \log \qa + \beta \log \qa - \beta \log \qa^{(j)} - \gamma \right) d\traj d\theta + \gamma \label{eqn:integral},
\end{equation}
where, we used the relations:
\begin{equation}
  \langle \nabla_{\qt}\mathcal{J}, \qt \rangle = \mathcal{J},
\end{equation}
\begin{equation}
  \operatorname{KL}(\qt||\qt^{(j)}) = \int \qt \log\frac{\qt}{\qt^{(j)}}d\traj d\theta =
%  \int \qt \log\frac{p(\state|\action, \theta)\modelpost \qa}{p(\state|\action, \theta)\modelpost \qa^{(j)}} d\traj d\theta =
  \int \qt \log\frac{\qa}{\qa^{(j)}} d\traj d\theta.
\end{equation}
The integrand of \eqref{eqn:integral} can be organized as
\begin{align}
\qt \cdot \log \frac{\qa^{\alpha + \beta}}{p(\optimal|\traj)e^{-\gamma}(\qa^{(j)})^{\beta}}
&\propto
\qt \cdot \log \frac{\qa}{p(\optimal|\traj)^{\frac{1}{\alpha + \beta}}\cdot e^{\frac{-\gamma}{\alpha + \beta}}\cdot (\qa^{(j)})^{\frac{\beta}{\alpha + \beta}}} \\
&= \qt \cdot \log \frac
{p(\state|\action, \theta) \modelpost  \qa}
% ---------------------------------------
{(p(\state|\action, \theta) \modelpost \qa^{(j)})\cdot p(\optimal|\traj)^{\frac{1}{\alpha + \beta}}\cdot e^{\frac{-\gamma}{\alpha + \beta}}\cdot (\qa^{(j)})^{\frac{-\alpha}{\alpha + \beta}}} \\
&=
\qt \cdot \log \frac
{\qt}
% ---------------------------------------
{\qt^{(j)} \cdot p(\optimal|\traj)^{\frac{1}{\alpha + \beta}}\cdot e^{\frac{-\gamma}{\alpha + \beta}}\cdot (\qa^{(j)})^{\frac{-\alpha}{\alpha + \beta}}}.
\end{align}
Integrating the above equation yields,
\begin{equation}
\eqref{eqn:integral} =
\operatorname{KL}(\qt||\qt^{(j)} \cdot {p(\optimal|\traj)^{\frac{1}{\alpha + \beta}}\cdot
 e^{\frac{-\gamma}{\alpha + \beta}}\cdot (\qa^{(j)})^{\frac{-\alpha}{\alpha + \beta}}}) + \gamma.
\end{equation}
By minimizing this equation, we get:
\begin{equation}
\qt^{(j+1)} = {\qt^{(j)}\cdot p(\optimal|\traj)^{\frac{1}{\alpha + \beta}}\cdot e^{\frac{-\gamma}{\alpha + \beta}}\cdot (\qa^{(j)})^{\frac{-\alpha}{\alpha + \beta}}}.
% =
% \qa^{(j)}\cdot {p(\optimal|\traj)^{\frac{1}{\alpha + \beta}}\cdot e^{\frac{-\gamma}{\alpha + \beta}}\cdot (\qa^{(j)})^{\frac{-\alpha}{\alpha + \beta}}}
\label{eqn:update_law_w_lagrange}
\end{equation}
The Lagrange multiplier can be removed using the constraint $\int\qt^{(j+1)}\cdot d\traj d\theta=1$:
\begin{align}
e^{\frac{\gamma}{\alpha + \beta}} &= \mathbb{E}_{\qt^{(j)}}\left[p(\optimal|\traj)^{\frac{1}{\alpha + \beta}}\cdot (\qa^{(j)})^{\frac{-\alpha}{\alpha + \beta}} \right] \\
&=
\mathbb{E}_{\action \sim \qa~{(j)}}\left[
\underbrace{
\mathbb{E}_{\state \sim p(\state|\action, \theta), \theta \sim \modelpost}
\left[
p(\optimal|\traj)^{\frac{1}{\alpha+\beta}}
\right]
}_{(*) \label{eqn:lagrange}
% \simeq
}
\cdot (\qa^{(j)})^{\frac{-\alpha}{\alpha + \beta}}
\right].
\end{align}
Considering the discussion in \secref{sec:control_inference} and \secref{sec:comp_w}, we compute $(*)$ as
\begin{equation}
(*) \simeq f(\mathbb{E}[r(\tau)])^{\frac{1}{\alpha + \beta}} = \mathcal{W}'(\action)^{\frac{1}{\alpha + \beta}}.
\end{equation}
Substituting \eqref{eqn:lagrange} to \eqref{eqn:update_law_w_lagrange} results in:
\begin{equation}
\qt^{(j+1)} = \frac
{\qt^{(j)}\cdot p(\optimal|\traj)^{\frac{1}{\alpha + \beta}}\cdot (\qa^{(j)})^{\frac{-\alpha}{\alpha + \beta}}}
% ---------------------------------------------------------------------------------------------------------------
{\mathbb{E}_{\action \sim \qa^{(j)}}\left[\mathcal{W}'(\action)^{\frac{1}{\alpha + \beta}}\cdot (\qa^{(j)})^{\frac{-\alpha}{\alpha + \beta}} \right]}.
\end{equation}
Marginalizing $(\state, \theta$), we finally obtain:
\begin{equation}
\qa^{(j+1)} = \frac
{\qa^{(j)}\cdot \mathcal{W}'(\action)^{\frac{1}{\alpha + \beta}}\cdot (\qa^{(j)})^{\frac{-\alpha}{\alpha + \beta}}}
% ---------------------------------------------------------------------------------------------------------------
{\mathbb{E}_{\action \sim \qa^{(j)}}\left[\mathcal{W}'(\action)^{\frac{1}{\alpha + \beta}}\cdot (\qa^{(j)})^{\frac{-\alpha}{\alpha + \beta}} \right]}.
\end{equation}
In \eqref{eqn:vi}, we replaced $\lambda := \alpha + \beta$, $\kappa := \alpha /(\alpha + \beta)$.

% \begin{equation}
% q^{(j+1)} = \frac{q^{(j)}\cdot p(\optimal|\traj)^{\frac{1}{\lambda}}\cdot q^{-\kappa}^{(j)}}{\mathbb{E}_{q^{(j)}}\left[p(\optimal|\traj)^{\frac{1}{\lambda}}\cdot q^{-\kappa}^{(j)} \right]}
% \end{equation}

% \begin{equation}
% q^{(j+1)} = \frac{q^{(j)} \exp\left\{\frac{1}{\alpha + \beta}\log p(\optimal|\traj) - \frac{\beta}{\alpha + \beta}\log q^{(j)}\right\}}{\mathbb{E}_{q^{(j)}}\left[\exp\left\{\frac{1}{\alpha + \beta}\log p(\optimal|\traj) - \frac{\beta}{\alpha + \beta}\log q^{(j)}\right\}\right]}
% \end{equation}

\section{Complete Definition of \proposedmethod{}} \label{sec:paets_eqs}
\begin{align}
% \begin{equation}
    \resp_{m}(\action_{k}) &:= %\frac
    {\mixcoef^{(j)}_{m}\mathcal{N}(\action_{k} ; \actionm^{(j)}_{m}, \actions^{(j)}_{m})} {\Big /}
    % -----------------------------------------------------------------------------
    {\sum^{M}_{m'=1}\mixcoef^{(j)}_{m'}\mathcal{N}(\action_{k} ; \actionm^{(j)}_{m'}, \actions^{(j)}_{m'})} \label{eqn:resp} \\
% \end{equation}
%
% \begin{equation}
    % N_{m} := {},\
    \omega_{m, k}^{(j+1)} &:= %\frac
        {\resp_{m}(\action_{k}) w_{k}^{(j+1)}} {\Big /}
        % ------------------------------
        {\underbrace{\sum^{K}_{k'=1} \resp_{m}(\action_{k'}) w_{k'}^{(j+1)}}_{:= N_{m}}} \\
% \end{equation}
%
% \begin{align}
    \actionm_{m}^{(j+1)} &\leftarrow
        {\sum^{K}_{k=1} \omega^{(j+1)}_{m,k}\action_{k}} \\
%
    % \actions_{m}^{(j+1)} &\leftarrow&
    %     {\sum^{K}_{k=1} \omega^{(j+1)}_{m,k}(\action_{k} - \actionm_{m}^{(j+1)})(\action_{k} - \actionm_{m}^{(j+1)})^{\mathrm{T}}} \\
    \actions_{m}^{(j+1)} &\leftarrow
        {\sum^{K}_{k=1} \omega^{(j+1)}_{m,k}(\action_{k} - \actionm_{m}^{(j+1)})^{2}} \\
    \mixcoef_{m}^{(j+1)} &\leftarrow N_{m} {\Big /} \sum^{M}_{m=1} N_{m}. \label{eqn:mix_coef}
\end{align}

\section{Optimization of Toy Objective Function by \proposedmethod{}} \label{sec:toytask_opt}
\begin{figure}[h]
  \centering
  \includegraphics[width=0.7\textwidth]{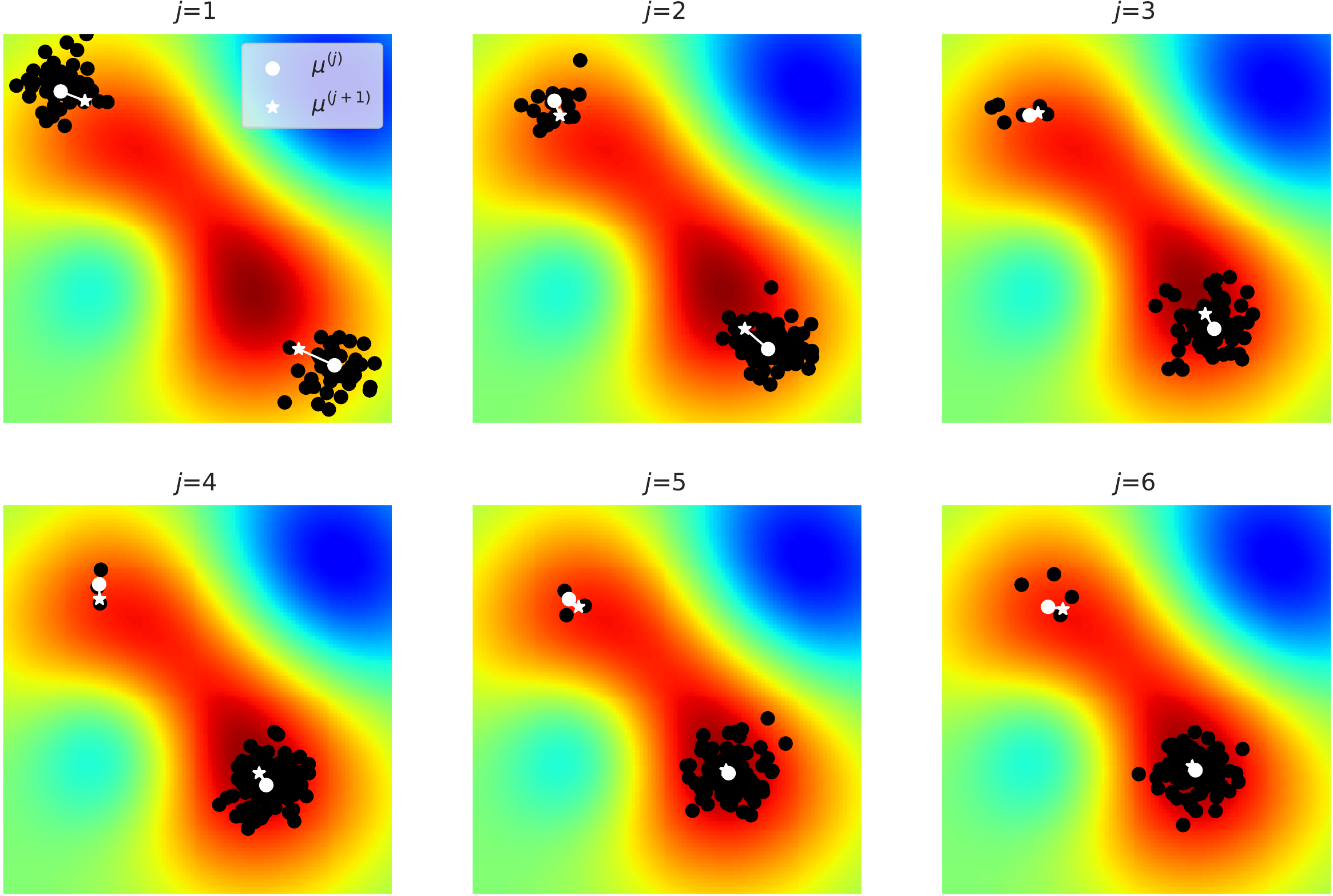}
  \caption{
  The optimization process of a 2D multimodal objective function by \proposedmethod{} (\texttt{VIMPC(`MPPI', `GMM(M=2)', True)}),
  in which two distribution components are successfully optimized to fit the two modals.
  $\bullet$ depict particles that approximates $q(\action; \phi^{(j)})$.
%   In this case, our method successfully captured the multimodality.
  }
  \label{fig:toytask_opt}
\end{figure}

Fig.~\ref{fig:toytask_opt} illustrates how \proposedmethod{} optimizes $q(\action; \phi^{(j)})$ in a toy multimodal objective function.

\section{Computational Complexity} \label{sec:comp_comp}
The main computational bottleneck of \proposedmethod{} (and PETS) is the execution of $\ell$3--6 in Alg.~\ref{alg:vi_traj_opt}, in which total $K\times P$ trajectories must be sampled. In our experiment, $K$ and $P$ were respectively set as $K=500$, $P=20$ as in~\citep{chua2018deep}.
Compared to PETS, \proposedmethod{} requires additional procedures like action sampling from GMM ($\ell$2) and GMM parameter update ($\ell$9).
However, these additional procedures are easily parallelizable on GPUs, and their computation times are much shorter than the above mentioned bottleneck.
In the experiments with our early prototype in TensorFlow, it took about 57~ms for $M=5$ and 55~ms for $M=1$ (equivalent to PETS) to execute one iteration of the for-loop in Alg.~\ref{alg:vi_traj_opt} on a single NVIDIA RTX2080 GPU.
% Although our implementation completely parallelizes $K\times P$ trajectory sampling on the GPU,
The above execution time does not meet the real-time constraints~(e.g., 30~Hz).
However, considering the success of the real-time implementation of MPPI in~\citep{williams2016aggressive,williams2017information},
we believe real-time implantation of our method is feasible with optimized implementation using compiled language, low-level GPU APIs,
and thorough tuning of hyperparameters (e.g., $K$, $P$ and DNN complexity).

\section{Implementation Notes} \label{sec:impl_notes}
\paragraph{Cross Entropy Method}
It is general technique to adaptively determine $r_{thd}$ in \tabref{tab:opt_algs}
so that only the top-$e\%$ samples satisfies the threshold condition.
We employ this technique and the \textit{eliteness} ratio is set to be $e=10\%$. $\lambda$ has no effect on CEM optimization since $f(\cdot)$ takes binary values.

\paragraph{MPPI}
Reward normalization heuristics, as suggested in~\citep{theodorou2010generalized}, were also introduced for our MPPI implementation as
\begin{equation}
  \mathcal{W}'(\action_{k})^{\frac{1}{\lambda}} = \exp\left\{
  \frac{1}{\lambda} \cdot
  \frac
  {r(\traj_{k}) - \operatorname{min}\{r(\traj_{k'}) \}_{k'=1}^{K}}
  % ---------------------------------------------------------------
  {\operatorname{max}\{r(\traj_{k'}) \}_{k'=1}^{K} - \operatorname{min}\{r(\traj_{k'}) \}_{k'=1}^{K}}
  \right\},
\end{equation}
where $r(\traj_{k}) = \frac{1}{P}\sum^{P}_{i=1}r(\traj_{k,i})$. $\lambda$ was set to be $\lambda = 0.1$ as also suggested in~\citep{theodorou2010generalized}.

\paragraph{Entropy Regularization}
The value of $\kappa$ is very sensitive to task settings, especially for the dimensionalities of action spaces.
To make $\kappa$ insensitive, we introduced the following normalization trick inspired by the above heuristics.
First, we rearrange \eqref{eqn:weight_update} as
\begin{equation}
  w^{(j+1)}_{k} \propto \mathcal{W}'(\action)^{\frac{1}{\lambda}} \exp\left\{\kappa \cdot (-\log q^{(j)}(\action_{k}))\right\}.
\end{equation}
Then, we replace $-\log q^{(j)}(\action_{k})$ to normalized one:
\begin{equation}
  -\log q^{(j)}(\action_{k}) \rightarrow
  \frac
  {-\log q^{(j)}(\action_{k}) - \operatorname{min}\{-\log q^{(j)}(\action_{k'})\}_{k'=1}^{K}}
  % ---------------------------------------------------------------------------------------
  {\operatorname{max}\{-\log q^{(j)}(\action_{k'})\}_{k'=1}^{K} - \operatorname{min}\{-\log q^{(j)}(\action_{k'})\}_{k'=1}^{K}} \in [0, 1].
\end{equation}
By applying these  heuristics, the range of entropy bonus is limited to $[1, e^{\kappa}]$,
where the action with the lowest probability among $K$ samples gains the highest entropy bonus of $e^{\kappa}$.
% while the action with highest probability gains no bonus $e^{0} = 1$.

\section{Experimental Setup} \label{sec:exp_setup}
We used MuJoCo tasks modified from standard OpenAI Gym tasks.%
\footnote
{\url{https://github.com/openai/gym}}
\tabref{tab:reward_func} summarizes the task settings, where $v_{x}$, $\varphi$ and $z$ respectively denote the velocity, orientation angle, and height of the agents.
Penalty functions $\Phi$, $\Psi$ are newly introduced to encourage the agents to move forward in the proper form.
Instead, \texttt{done} flags used originally for early task stopping are removed.
$\Phi$, $\Psi$ are defined as
\begin{equation}
  \Phi(z, z_{des}) = e^{-(z - z_{des})^{2}},
\end{equation}
\begin{equation}
  \Psi(\varphi) = \frac{1 + \cos(2\varphi)}{2}.
\end{equation}

We modified the range of actions (i.e.,~torques) from $[-1, 1]$ to $[-5, 5]$
to exaggerate uncertainties in the optimal trajectory posteriors.
\begin{table}[h]
  \caption{MuJoCo task settings.}  \label{tab:reward_func}
  \centering
  \begin{tabular}{c|c|c|c|c}
    Task & Reward Function & $\state_{t} \in $ & $\action_{t} \in $ & Misc. \\\hline\hline
    HalfCheetah & $v_{x} \cdot \frac{1 + \operatorname{sign}(\cos(\varphi))}{2} - 0.1 \cdot ||\action_{t}||^{2}$
    & $\mathbb{R}^{18}$ & $\mathbb{R}^{6}$ & --- \\
    Ant & $v_{x} \cdot \Phi(z, z_{des}) - 10^{-3} \cdot ||\action_{t}||^{2}$
    & $\mathbb{R}^{28}$ & $\mathbb{R}^{8}$ & $z_{des} = 0.75$ \\
    Hopper & $v_{x} \cdot \Phi(z, z_{des}) \cdot \Psi(\varphi) - 10^{-3} \cdot ||\action_{t}||^{2}$
    & $\mathbb{R}^{12}$ & $\mathbb{R}^{3}$ & $z_{des} = 1.2$ \\
    Walker2d & $v_{x} \cdot \Phi(z, z_{des}) \cdot \Psi(\varphi) - 10^{-3} \cdot ||\action_{t}||^{2}$
    & $\mathbb{R}^{18}$ & $\mathbb{R}^{6}$ & $z_{des} = 1.2$ \\
  \end{tabular}
\end{table}

\tabref{tab:mbrl_params} summarizes the shared parameter settings for MBRL (\proposedmethod{}, PETS, and MPPI).
For SAC, we used the default parameters from the original codebase.
\begin{table}[h]
  \caption{MBRL parameters.}  \label{tab:mbrl_params}
  \centering
  \begin{tabular}{c|cccc}
    & HalfCheetah & Ant & Hopper & Walker2d \\\hline\hline
    $T$: prediction horizon & 30 & 30 & 60 & 45 \\
    $\kappa$: weight of entropy regularizer & 0.5 & 0.25 & 0.5 & 0.5 \\
    $K$: \# sampled actions & \multicolumn{4}{c}{500} \\
    $P$: \# trajectories for each action & \multicolumn{4}{c}{20} \\
    $U$: \# optimization-iterations & \multicolumn{4}{c}{5} \\
    $H$: \# episode length & \multicolumn{4}{c}{1000} \\\hline
    $E$: \# neural networks & \multicolumn{4}{c}{5} \\
    hidden nodes & \multicolumn{4}{c}{(200, 200, 200, 200)} \\
    activation function & \multicolumn{4}{c}{Swish} \\
    optimizer & \multicolumn{4}{c}{Adam} \\
    learning rate & \multicolumn{4}{c}{$10^{-3}$} \\
    batch-size & \multicolumn{4}{c}{160} \\
  \end{tabular}
\end{table}

\section{Diversity Analysis of $\mathcal{D}$} \label{sec:diversity}
\begin{figure}[h]
  \centering
  \includegraphics[width=\textwidth]{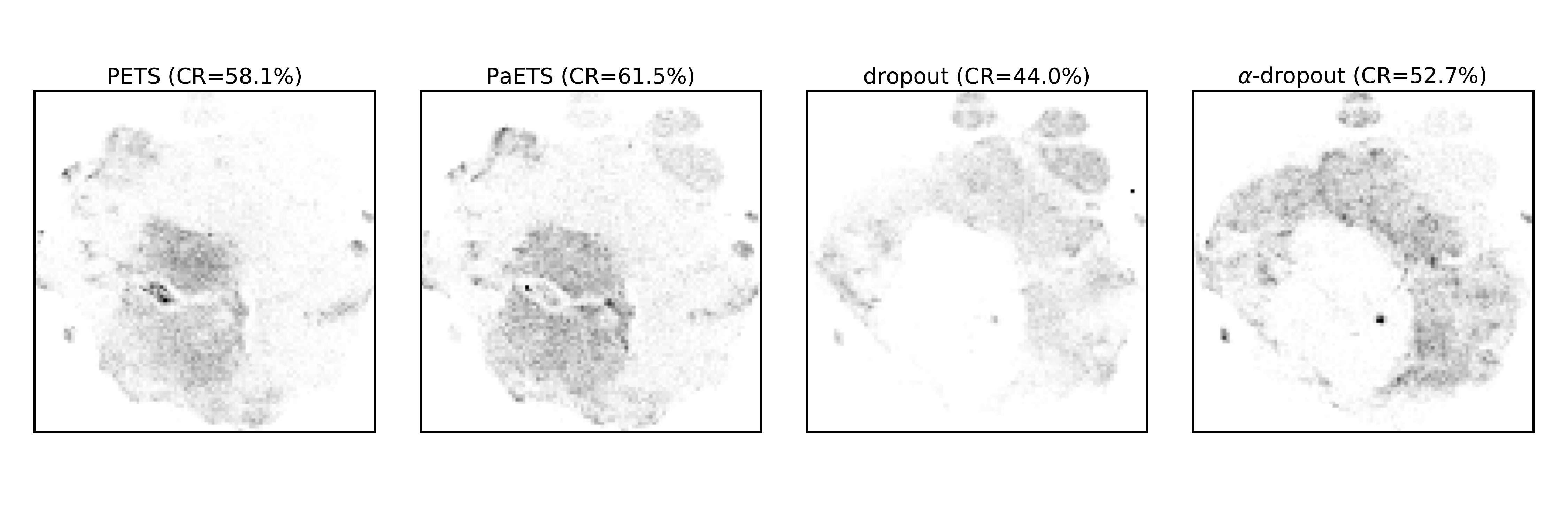}
  \caption{Comparison of training data distributions collected by different MPC-policies.
  CR (cover ratio) indicates the ratio of non-zero bins in each 2D histogram.}
  \label{fig:coverage}
\end{figure}

In this section, we analyzes the diversity of training data $\mathcal{D}$ collected by different MPC-policies.
The distributions (histograms) of the data samples are illustrated in \figref{fig:coverage},
in which the dimension of a sample ($\state$, $\action$) was reduced by t-SNE.
This figure suggests that incorporating uncertainty both in the dynamics and optimalities can improve the diversity of $\mathcal{D}$ (i.e.,~coverage of state-action space).

\end{document}